%% file: main_rev02.tex
\documentclass{article}
\usepackage[margin=1in]{geometry}
\usepackage{amsmath,amssymb}
\usepackage{graphicx} % Required for inserting images
\usepackage{xcolor}
\usepackage{subcaption}
\usepackage[colorlinks=true,linkcolor=blue,citecolor=blue]{hyperref}
\usepackage{multirow}
\usepackage{makecell}
\usepackage{bm}
\usepackage[square,numbers]{natbib}
\usepackage{tikz}
\usetikzlibrary{arrows.meta,positioning,shadows.blur,shapes.geometric,fit,calc}

\newcommand{\rev}[1]{%
  {\color{black} #1}%
}

\definecolor{basecolor}{RGB}{180,180,220}
\definecolor{targetcolor}{RGB}{180,180,220}
\definecolor{transformercolor}{RGB}{100,150,200}
\definecolor{arrowcolor}{RGB}{100,100,100}
\definecolor{textcolor}{RGB}{40,40,40}
\definecolor{inputcolor}{RGB}{200,100,100}
\definecolor{outputcolor}{RGB}{100,150,200}
\definecolor{greencolor}{RGB}{150,180,100}

\title{Conditional Normalizing Flows for Forward and Backward Joint State and Parameter Estimation}
\author{Luke~S.~Lagunowich$^1$, Guoxiang~Grayson~Tong$^2$, Daniele~E.~Schiavazzi$^3$}
\date{$^1$Department of Computer Science and Engineering\\
University of Notre Dame, Notre Dame, IN, 46556, USA\\
\vspace{3pt}
$^2$Department of Pediatrics\\
Stanford University, Stanford, CA, 94305, USA\\
\vspace{5pt}
$^3$Department of Applied and Computational Mathematics and Statistics\\
\vspace{3pt}
University of Notre Dame, Notre Dame, IN, 46556, USA
}

\begin{document}

\maketitle

%\tableofcontents

\begin{abstract}
\noindent Traditional filtering algorithms for state estimation -- such as classical Kalman filtering, unscented Kalman filtering, and particle filters -- show performance degradation when applied to nonlinear systems whose uncertainty follows arbitrary non-Gaussian, and potentially multi-modal distributions.
This study reviews recent approaches to state estimation via nonlinear filtering based on conditional normalizing flows, where the conditional embedding is generated by standard MLP architectures, transformers or selective state-space models (like Mamba-SSM).
In addition, we test the effectiveness of an optimal-transport-inspired kinetic loss term in mitigating overparameterization in flows consisting of a large collection of transformations.
We investigate the performance of these approaches on applications relevant to autonomous driving and patient population dynamics, paying special attention to how they handle time inversion and chained predictions.
Finally, we assess the performance of various conditioning strategies for an application to real-world COVID-19 joint SIR system forecasting and parameter estimation.
\end{abstract}

% ====================
\section{Introduction}
% ====================
\rev{The problem of estimating the hidden state of a dynamical system arises across a broad spectrum of applications in science and engineering.}
In robotics, autonomous vehicles estimate their state from sensor measurements, which are often stochastic, resulting in uncertainty about the true underlying state.
Similarly, in epidemiology, health agencies can infer the dynamics of underlying disease progression from uncertain diagnostic measurements, such as positive test cases.
In essence, \emph{forward state estimation} refers to the ability to infer the distribution of a $d$-dimensional state vector $\bm{x}_{n}\in\mathbb{R}^{d}$ at discrete times $t=n\Delta t$ with $n\in\{0,1,\dots,N\}$, given the $R$ observations $\{\bm{o}_{n-R},\bm{o}_{n-R+1},\dots,\bm{o}_{n-1}\}$, where $\bm{o}_{n-r}\in\mathbb{R}^{m}$, for $r=0,\dots,R$. 
Similarly, state estimation can be reversed \rev{(a process we refer to as \emph{backward state estimation})}, focusing on inference of a past state from future observations, or inferring $\bm{x}_{n}$ given the $R$ observations $\{\bm{o}_{n+1},\bm{o}_{n+2},\dots,\bm{o}_{n+R}\}$.
This can be useful for capturing the dynamics of a system given an incomplete set of past or future observations in time.
\rev{
For $R=n-1$ forward estimation is commonly referred to as \emph{filtering} in the literature. Similarly, \emph{smoothing} involves estimating $\bm{x}_{n}$ from the entire set of observations $\{\bm{o}_{0},\dots,\bm{o}_{N}\}$. Our unorthodox use of the terms \emph{forward} and \emph{backward} estimation in place of \emph{filtering} and \emph{smoothing}, is justified in light of the flexibility in the choice of $R$.
}
Additionally, while realistic decision-making processes -- such as those in autonomous driving or epidemiological forecasting -- often involve equally probable alternatives, classical state estimation techniques frequently struggle to capture complex multimodal distributions.
\rev{
This includes methods such as the Kalman filter~\cite{kalmanfilter}, which typically suffer performance degradation when applied to nonlinear systems or non-Gaussian noise. The Unscented Kalman filter (UKF~\cite{julier97,wan2000unscented}) and the Ensemble Kalman filter (EnKF, see, e.g.,~\cite{katzfuss2016understanding}) were introduced to relax the linearity requirement, yet both retain an underlying Gaussianity assumption. Particle filters~\cite{gordon1993novel}, by contrast, offer a flexible framework for nonlinear and non-Gaussian filtering, with extensions supporting joint state and parameter estimation~\cite{evensen2022data}; however, their convergence is known to deteriorate in high-dimensional and multimodal settings~\cite{bengtsson2008curse, Vaswani08}. Sequential Monte Carlo (SMC) algorithms~\cite{doucet2001introduction, chopin2013smc2} provide a more general framework for propagating and reweighting a population of samples, and can be viewed as a broad generalization of particle filters. For an in-depth review of Bayesian filtering and smoothing, the interested reader is referred to, e.g.,~\cite{sarkka2023bayesian}.

Rooted in optimal transport theory~\cite{villani2009optimal}, the Feedback Particle Filter (FPF~\cite{taghvaei2021optimal}) drives each particle toward the posterior distribution via a feedback control input that continuously incorporates new observations, thereby mitigating the weight degeneracy observed in traditional particle filters, but incurring additional cost per time step. A discrete-time formulation and an amortized counterpart of FPF are introduced in~\cite{al2023optimal} and~\cite{al2025fast}, respectively. The measure-theoretic neural mapping-enhanced ensemble filter (MNMEF~\cite{bach2025learning}) represents a further recent development, employing set transformers to transport forecast ensembles toward the posterior distribution. Finally, methods based on tensor train decomposition have also been proposed for sequential state and parameter estimation~\cite{zhao2024tensor}.
}

% Density estimation with deep neural networks
Recent developments combining density estimation and deep neural networks have offered new possibilities for \rev{inference of} non-Gaussian distributions for states or parameters in nonlinear models. 
% MDN
One class of such models are Mixture Density Networks (MDNs~\cite{bishopMDN}) which parametrize the linear combination of Gaussian kernel functions. MDNs have been applied in trajectory modeling~\cite{girbau2021multipleobjecttrackingmixture} and in rider demand forecasting as part of a novel recurrent architecture~\cite{li2024xrmdnextendedrecurrentmixture}. However, despite their increased performance with nonlinear systems, in their basic form MDNs struggle with multi-modal, non-Gaussian target distributions as discussed in~\cite{DBLP:journals/corr/abs-1809-02129}. 
% GANs
Generative Adversarial Networks (GANs~\cite{goodfellow2014generativeadversarialnetworks}) are a class of generative models that imitate complex distributions. However, GANs do not provide direct access to the generator distribution, preventing this approach to be used for density estimation and uncertainty quantification. 
% VAE
Variational Autoencoders (VAEs~\cite{kingma2022autoencodingvariationalbayes}) use an encoder-decoder structure to perform probabilistic generative modeling tasks by optimizing the encoding of a simple prior distribution to represent the posterior and estimating the decoding of such representation. VAEs have been used in applications of finance~\cite{vae_fin_bio}, speech processing~\cite{vae_speech}, and bio-signal processing~\cite{Chen18} for their strength in learning complex input distributions for these tasks.

% Normalizing flow to model complex distributions
In this study we instead focus on Normalizing Flows (NF), which generate expressive probability distributions by applying a series of parametrized transformations \rev{to an easy-to-sample} \emph{base} density.
NF are traditionally used in density estimation and generative modeling~\cite{papamakarios2021normalizingflowsprobabilisticmodeling}, and have been used to model complex distributions in time-series and image modeling~\cite{papamakarios2021normalizingflowsprobabilisticmodeling}, image generation~\cite{ho2019flowimprovingflowbasedgenerative}, noise modeling~\cite{Abdelhamed_2019_ICCV}, and physics~\cite{Kanwar_2020, köhler2019equivariantflowssamplingconfigurations, doi:10.1126/science.aaw1147, Wirnsberger_2020, li2020scalablegradientsstochasticdifferential}. 
\rev{
Recent developments include differentiable particle filters based on normalizing flows~\cite{chen2024normalizing}. Closely related ideas combining continuous normalizing flows with optimal transport costs -- enforced through satisfaction of the Hamilton–Jacobi–Bellman equations -- are explored in~\cite{onken2021ot}, with extensions to flow matching introduced in~\cite{kerrigan2024dynamic}. Approaches based on optimal transport maps, conceptually related to normalizing flows, have also been proposed and applied to both the filtering and smoothing problems in~\cite{el2012bayesian, spantini2022coupling, Grashorn2023, ramgraber2023ensemble}.
}
Most significant to our work is the paper developed by Delecki et al.~\cite{delecki2023deepnormalizingflowsstate} which combines transformers with conditional normalizing flows.
% Surrogate modeling
Finally, estimation approaches involving particles may require the repeated solution of an expensive physics-based solver leading to computational intractability. This can be mitigated with approaches where the underlying deterministic solver is replaced by an inexpensive  surrogate~\cite{queipo2005surrogate,marzouk2007stochastic}.

% What we propose
We propose various normalizing flow architectures parametrized with transformer and selective state-space model-based conditioning operators.
Note that recurrent conditioning operators can generate embedding from sequences of varying length, providing unparalleled flexibility for filtering problems with an arbitrary number of past or future observations. We aim to combine the flexibility of these conditioning mechanisms with the expressive power of normalizing flows in representing a large class of densities for state-only and joint state and parameter estimation under uncertainty.

The paper is organized as follows. In Section~\ref{sec:dynamical_systems}, we start by introducing two dynamical systems of interest with applications in autonomous driving and epidemiology. Section~\ref{sec:nf} introduces normalizing flow and conditional normalizing flow architectures. Various strategies to generate conditional embeddings are discussed in Section~\ref{sec:cond_embedding}, focusing on transformers and selective state-space models. 
In Section~\ref{sec:kinetic} we also introduce a kinetic term in the loss function and assess its effectiveness on density estimation accuracy. 
Results in Section~\ref{sec:results} show the performance of the proposed approaches on the two selected dynamical systems, including forecast with real data from the COVID-19 pandemic. 
A discussion with possible ideas for future work are finally provided in Section \ref{sec:discussion}.

% ==================================
\section{Methods}\label{sec:methods}
% ==================================

% =========================================================
\subsection{Dynamical Systems}\label{sec:dynamical_systems}
% =========================================================

To validate the proposed architectures, we introduce two dynamical systems in the next sections.

% ===============================================================
\subsubsection{Autonomous Vehicle Dynamics with Random Switching}\label{sec:ds_driving}
% ===============================================================

We first consider a discrete bimodal dynamical system given by a set of four difference equations with a random switch parameter, originally proposed in~\cite{delecki2023deepnormalizingflowsstate}.
This model represents the motion of an autonomous vehicle, where the bimodal trajectory could result from sensor readings while traversing a roundabout or intersection or after a sudden change in trajectory.
To simulate realistic sensor data, we add noise to the trajectory determined by the following equations of motion. 
Consider the two-dimensional trajectory of a vehicle with location at time $t$ expressed by the pair $(p^{t}_x, p^{t}_y)$ and heading angle $\theta^{t}$, which is updated at every time step according to the equations 
\begin{equation}
\begin{split}
p_x^{t+1} &= p_x^t + \Delta t \cdot v^t \cdot cos(\theta^t)\\
p_y^{t+1} &= p_y^t + \Delta t \cdot v^t \cdot sin(\theta^t)\\
\theta^{t+1} &= \theta^t + \Delta t\cdot v^t \cdot \phi^t,
\end{split}
\end{equation}
where the angular acceleration $\phi^t$ at time $t$ is updated with the formula
\begin{equation}
\phi^{t+1} = \phi^t + \Delta t \cdot \psi \cdot c_1 \cdot cos(c_2 \cdot t).
\end{equation}
The autonomous vehicle is nonholonomic and has control over its velocity $v^t$ and angular acceleration $\phi^t$. Uncertainty in vehicle position accumulates over time as Gaussian noise is added to $v^t$ and $\phi^t$ with $\sigma_v=0.01$ and $\sigma_{\phi} =\rev{0.001}$, respectively, at each time step. The quantities $c_1=0.1$ and $c_2=0.5$ are constants, and $\psi$ is a random switching parameter which is drawn from a uniform distribution $\psi\sim\textit{U}([-1,1])$ at a fixed time index $t=5.5$ in the trajectory. This parameter is used to introduce multi-modality in the resulting trajectory.

\rev{Three datasets are generated using these dynamics. First, a full dataset with a fixed time step of $\Delta t = 1.0$ is selected for comparison with baseline methods and is shown in Figure~\ref{fig:autonomous_data_a}. Second, a full dataset with random dropout resulting in varying time steps consisting of approximately of 1.5 million data points for comparability to~\cite{delecki2023deepnormalizingflowsstate} is selected and is shown in Figure~\ref{fig:autonomous_data_b}. Third, a sparse dataset with random dropout resulting in varying time steps and consisting of approximately $1,500$ data points is selected to show the robustness of our methods and is shown in Figure~\ref{fig:autonomous_data_c}. In each figure, the black line represents the nominal noiseless trajectory.}

% ===================================================================
\subsubsection{Epidemiological Population Dynamics}\label{sec:ds_sir}
% ===================================================================

As a second model, we consider a simple compartmental model to represent the susceptible, infectious, and removed, by death or by recovery, (S, I, and R) patient populations with applications to the epidemiology of an infectious disease.
We purposefully omit general birth and death rates from our model, assuming the dynamics of an epidemic are much more rapid than the dynamics of birth and death. 
The SIR model can be represented by the following system of ordinary differential equations 
\begin{equation}
\frac{dS}{dt} = 
\dot{S}= - \beta\,I\,S,\,\,\frac{dI}{dt} = \dot{I} = \beta\,I\,S - \gamma\,I,\,\,\frac{dR}{dt} = \dot{R} = \gamma\,I,
\end{equation}
where $S$ is the number of susceptible individuals, $I$ the number of infected individuals, and $R$ the number of removed (dead or recovered) individuals. 
In addition, $\beta$ is the infection rate, and $\gamma$ is the recovery rate. For the simulated model, we select the parameters $\beta=0.03$ and  $\gamma=0.01$ to ensure the reproduction number, the number of new infected individuals an infected person can cause, $R_0>1$, yielding nontrivial epidemic dynamics over a moderate time horizon. The system is solved using the Fourth Order Runge-Kutta method. Noise with $\sigma=\rev{0.027}$ is \rev{added to observations from each trajectory in the system}. The low magnitude is chosen to improve robustness without affecting the dynamics of the underlying system.
\rev{Figure~\ref{fig:sir_data}} shows simulated noisy trajectory observations for each of the three sub-populations, where the black lines represent the nominal trajectories. Similar to the previous application, this dataset contains over 1.5 million data points.
\rev{
\begin{figure}[!ht]
    \centering
    \begin{subfigure}[b]{0.32\textwidth}
        \centering
        \includegraphics[width=\textwidth]{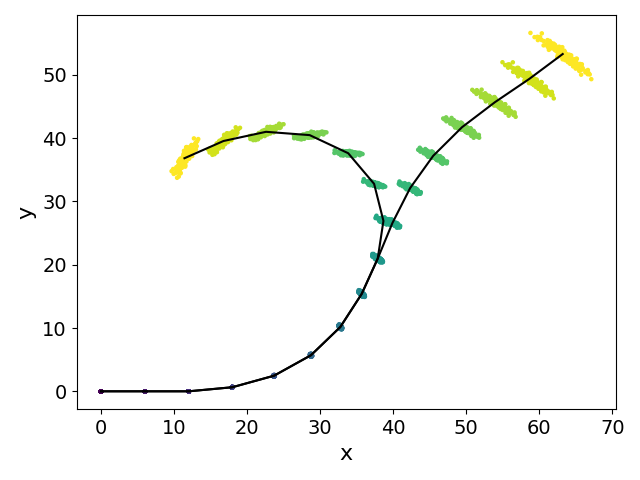}
        \caption{}
        \label{fig:autonomous_data_a}
    \end{subfigure}
    \hfill
    \begin{subfigure}[b]{0.32\textwidth}
        \centering
        \includegraphics[trim={0pt -6pt 0pt 0pt}, clip,width=\textwidth]{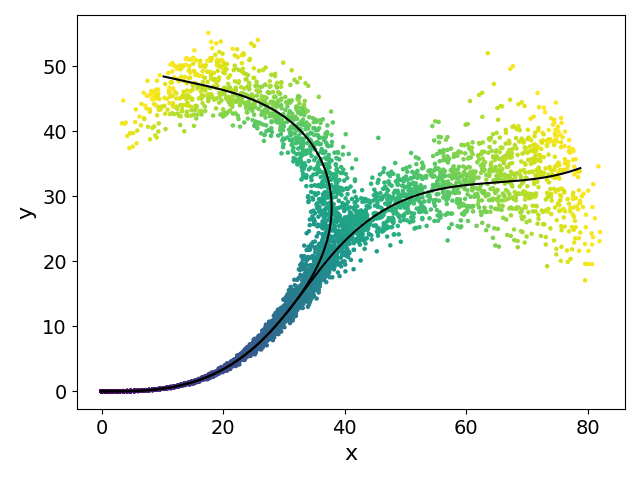}
        \caption{}
        \label{fig:autonomous_data_b}
    \end{subfigure}
    \begin{subfigure}[b]{0.32\textwidth}
        \centering
        \includegraphics[trim={0pt -6pt 0pt 0pt}, clip,width=\textwidth]{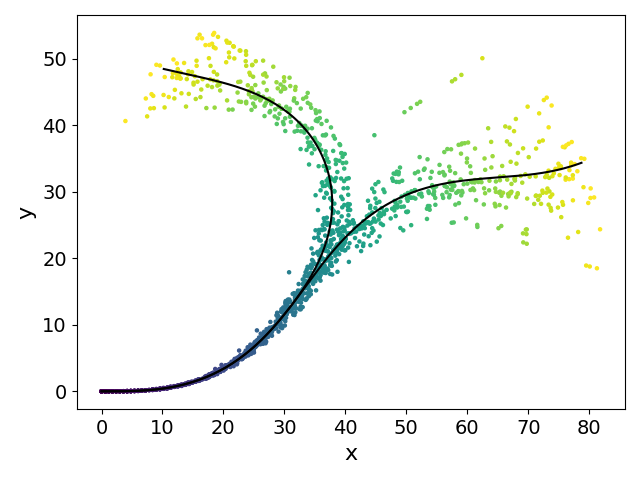}
        \caption{}
        \label{fig:autonomous_data_c}
    \end{subfigure}
    \caption{Sensor data for autonomous vehicle with random switching (a) full with fixed time steps, (b) full with random dropout, and (c) sparse with random dropout.}
    \label{fig:autonomous_data}
\end{figure}

 \begin{figure}[!ht]
     \centering
     \includegraphics[width=0.65\linewidth]{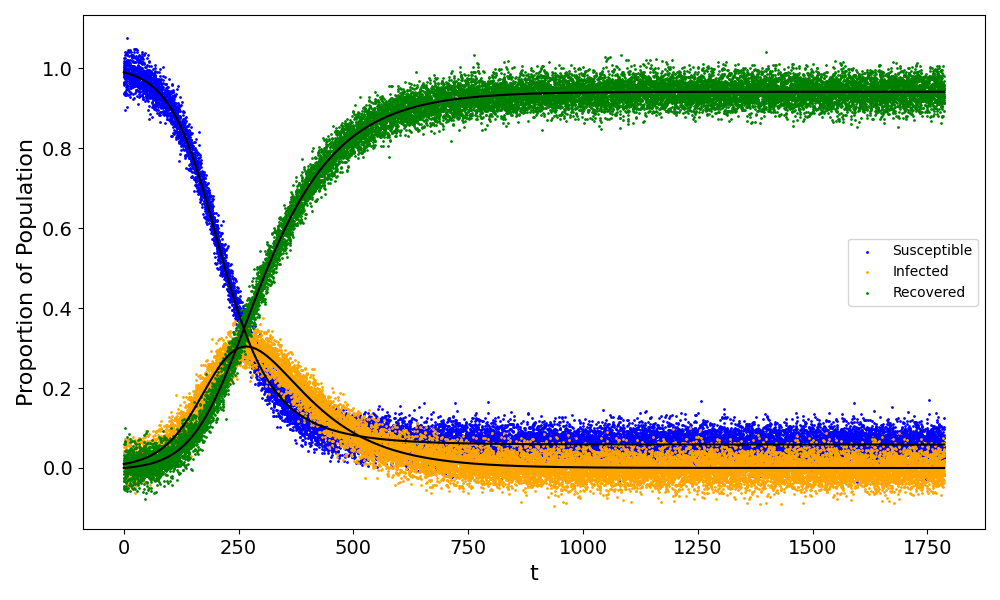}
     \caption{Noisy trajectories from the dynamical epidemiological SIR model.}
     \label{fig:sir_data}
 \end{figure}
}

% ===============================================================
\rev{\subsection{Baselines and Comparative Methods}\label{sec:baselines}
% ===============================================================

% ===========================================
\subsubsection{Particle Filter}\label{sec:pf}
% ===========================================

To establish a baseline comparison for the proposed methods, we implement a bootstrap particle filter~\cite{doi:10.1049/ip-f-2.1993.0015, doucet2001introduction}, as described in~\cite{sarkka2013bayesian}. 
The algorithm uses a set of $N$ particles $\{ (w_n^{(i)}, {\bm{x}}_n^{(i)})\}_{i=1}^{N}$ to represent the posterior distribution $p(\bm{x}_{n} | \bm{o}_{1},\dots,\bm{o}_{n})$. 
First, samples are drawn from the prior \(x^{(i)}_0 \sim p(x_0)\) and weights set to \(w^{(i)}_0 = \frac{1}{N}\).
Then, for each subsequent iteration, samples are drawn from the dynamic system \( p(\bm{x}_{n} | \bm{x}_{n-1})\) to update the population $\{ \bm{x}^{(i)}_{n-1}\}_{i=1}^{N}$. 
Weights are updated according to $w_n^{(i)} \propto p(\bm{o}_n | \bm{x}^{(i)}_n)$, triggering particle resampling whenever the effective sample size $n_{\text{eff}}$ drops below $N/2$.
The implementation uses systematic resampling~\cite{carpenter1999improved}. We implement the bootstrap particle filter, which we refer to and is known as the sequential Monte Carlo (SMC) algorithm, with $N= 1,000$ particles to estimate the posterior distribution $p({\bm{x}}_{n+4} | \bm{o}_{n+1},\dots,\bm{o}_{n+4})$ with $p(\bm{o}_n)$ as the prior, using four steps of the algorithm described above. We utilize the implementation provided through the {\it pyro} Python library~\cite{bingham2019pyro}.  
}

% ===============================================================
\subsection{Conditional Density Estimation with Normalizing Flow}\label{sec:nf}
% ===============================================================

In this study, we explore state and parameter estimation using a normalizing flow-based architecture, where we assess the performance of various conditioning operators and loss function formulations.

% General introduction to NF
A normalizing flow (NF) consists of a series of invertible mappings that characterize the transformation of a probability density. In more detail, a NF maps a simple \emph{latent} distribution $p_{\bm{Z}}(\bm{z})$ to an arbitrary target density $p_{\bm{X}}(\bm{x})$ through a collection of invertible transformations via the change of variable formula 
\begin{equation}
p_{\bm{X}}(\bm{x}) = p_{\bm{Z}} (f^{-1}(\bm{x})) \left|\det\left(\frac{d f^{-1} (\bm{x})}{d \bm{x}}\right)\right|,
\end{equation}
where $\bm{f}(\cdot,\bm{\theta}) = \bm{f}_1(\cdot,\bm{\theta}_{1}) \circ \bm{f}_2(\cdot,\bm{\theta}_{2}) \circ ... \circ \bm{f}_L(\cdot,\bm{\theta}_{L})$ is a composition of $L$ smooth, invertible diffeomorphisms, and $\bm{\theta}=\{\bm{\theta}_{1}, \bm{\theta}_{2},\dots,\bm{\theta}_{L}\}$. 
NF is a widely used machine learning paradigm for density estimation. Once trained, it allows for both the efficient generation of new samples and likelihood evaluation. This requires each transformation $f_i,\,i=1,\dots,L$ to be easily invertible, and fast computation for the determinant of its Jacobian.
% Various flows
Various formulations are proposed in the literature and the interested reader is referred to the two reviews in~\cite{kobyzev2020normalizing, papamakarios2021normalizingflowsprobabilisticmodeling} for additional details. 
% RealNVP and MAF 
For discrete NF, two widely popular approaches are based on transformations consisting of \emph{affine couplings}, for example RealNVP~\cite{dinh2016density},  or \emph{autoregressive transformations}, like masked autoregressive flows (MAF) introduced in~\cite{papamakarios2017masked}. 
% We use MAF
We use MAF in all examples in this study. Because in practice an autoregressive flow depends on the order of input variables, is has been proven beneficial to permute inputs between layers~\cite{papamakarios2017masked}. Thus the resulting flow architecture consists of stacked layers containing a permutation, a linear layer, and a masked affine autoregressive flow. 

% General idea of conditional NF
Importantly, NF architectures can also be used to generate samples from or evaluate the likelihood of conditional distributions. In practice, conditioning is incorporated into both the base distribution and each MAF layer. A conditioned base distribution allows the latent variables $\bm{z}$ to depend on the context. Similarly, within each MAF layer, the input is augmented so that the transformation variables also depend on the context.
We use the \textit{nflows} Python library~\cite{nflows} throughout this study, where a \textit{context} vector is injected into both the base distribution and each layer of the flow, consistent with the above discussion.

% ======================================================================
\subsection{Generating Conditional Embeddings}\label{sec:cond_embedding}
% ======================================================================

% Conditional probabilities to be estimated
We use normalizing flows to perform forward and \rev{backward} state estimation based on simulated observations from a dynamical system.
% Present from the past and from the future
For this task, we condition a normalizing flow on previous $\{\bm{o}_{n-R}, \bm{o}_{n-R+1}, \dots, \bm{o}_{n-1}\}$ and future $\{\bm{o}_{n+1}, \bm{o}_{n+2}, \dots, \bm{o}_{n+R}\}$ observations, taken from the noisy solutions of two dynamical systems. 
We then estimate the probability density of state $\bm{x}_{n}$, by approximating the probability distributions $p(\bm{x}_n | \bm{o}_{n-R}, \bm{o}_{n-R+1},\dots, \bm{o}_{n-1})$ and $p(\bm{x}_n | \bm{o}_{n+1}, \bm{o}_{n+2}, \dots, \bm{o}_{n+R})$, i.e., conditioned on $R$ past or future observations.
% This can be used recursively for rollout
Due to the sequential nature of time series data, we can utilize this conditioning operation recursively, estimating future states and then using these predicted states for conditioning, and so on and so forth. 
We call this paradigm \emph{rollout}, which allows to estimate all future (past) states of the system from a limited number of observations. 

\begin{figure}[!ht]
    \centering
    \begin{subfigure}[c]{0.34\textwidth}
        \centering
        \scalebox{0.7}{\input{fig_3a}}
        \caption{Normalizing flow with transformer conditioning operator.}
        \label{fig:nf_cond_transformer}
    \end{subfigure}
    \begin{subfigure}[c]{0.6\textwidth}
        \centering
        \scalebox{0.7}{\input{fig_3b}}
        \caption{Mamba-SSM diagram.}
        \label{fig:nf_cond_mamba}
    \end{subfigure}
    \caption{Generation of conditional embeddings using either a transformer \rev{or} Mamba-SSM architecture.}
    \label{fig:embedding}
\end{figure}
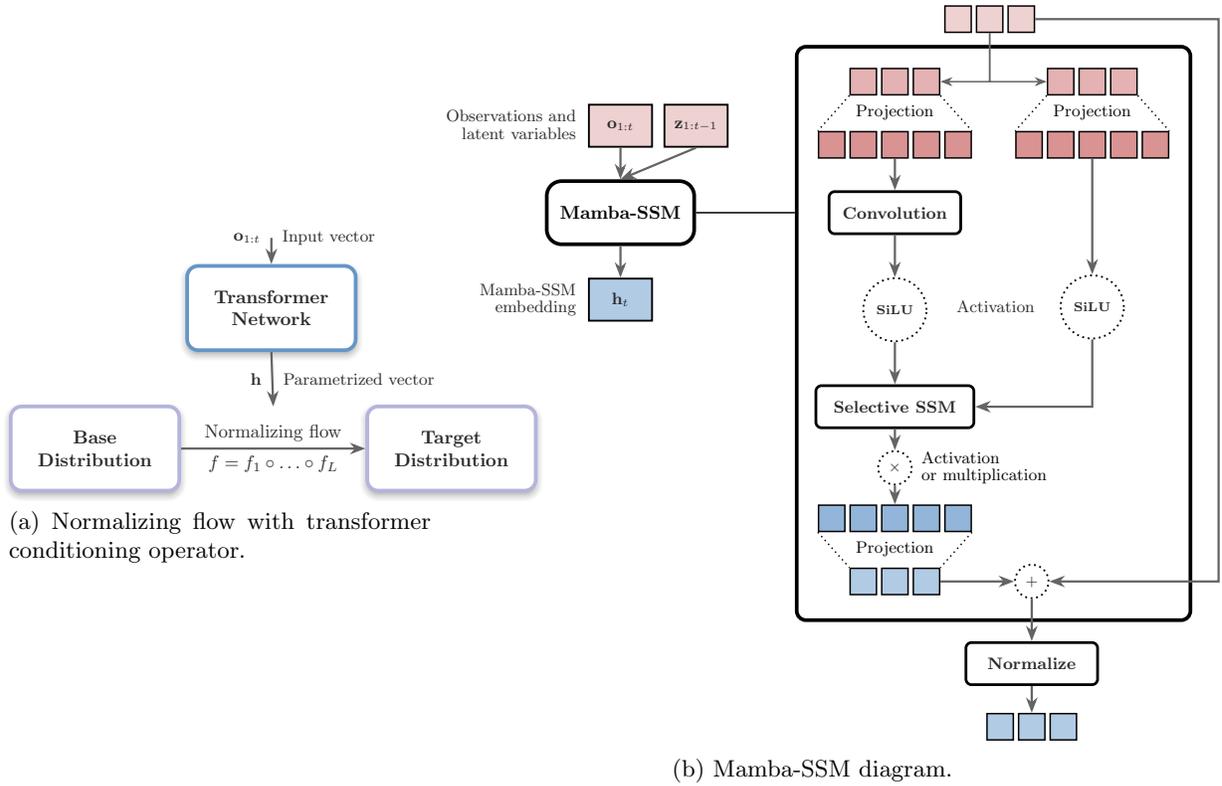

% Transformers
We consider two approaches to generate meaningful conditional embeddings. 
First, we use a transformer network to condition the flow on the provided context. A schematic of this idea is shown in Figure~\ref{fig:nf_cond_transformer}. We employ a model architecture like the one described in~\cite{wu2020deeptransformermodelstime}. The model consists of encoder and decoder layers. The encoder is composed of an input layer, a positional encoding layer, and a stack of four identical encoder layers.  Each encoder layer consists of two sub-layers: a self-attention sub-layer and a fully-connected feed-forward sub-layer. The decoder is composed of an input layer, four identical decoder layers, and an output layer. In addition to the two sub-layers in each encoder layer, the decoder inserts a third sub-layer to apply self-attention mechanisms over the encoder output. We employ a one-position offset and look-ahead masking to prevent look-ahead bias. 
Instead of using the transformer architecture for sequence-to-sequence forecasting, we use it to learn an embedding. Since there is no ground truth target sequence in this approach, we can freely choose the dimensionality of the embedding space without being constrained by the output sequence structure. \rev{We use a transformer implementation in Python from~\cite{delecki2023deepnormalizingflowsstate}.}

As an alternative to transformers, state space models (SSMs) can be used to process sequential information. SSMs represent the hidden or internal dynamics of a system through a set of first-order recurrence relations. At each time step, the hidden state evolves according to a linear \emph{update} equation while outputs are generated through a \emph{measurement} equation. Both equations are typically first-order differential equations. 
More specifically, the system of equations defines a map between inputs $\bm{x}(t)$ and outputs $\bm{y}(t)$ through a hidden state $\bm{h}(t)$ through the equations
\begin{equation}
\begin{split}
    \bm{h}'(t)&=\bm{A}\,\bm{h}(t)+\bm{B}\,\bm{x}(t)\\
    \bm{y}(t)&=\bm{C}\,\bm{h}(t),
\end{split}
\end{equation}
where $\bm{A}$, $\bm{B}$, $\bm{C}$ are learnable matrices. 
$\bm{A}$ is the state matrix, $\bm{B}$ is the input or control matrix, and $\bm{C}$ is the output matrix. The model is then computed in two stages. First, the continuous system parameters are converted to discrete parameters using \emph{zero-order hold} (ZOH, see, e.g.~\cite{pohlmann2000principles}). Next, the model can be computed in two ways: linear recurrence or global convolution, each with implications in terms of both accuracy and efficient implementation. 
However, vanilla state space models are time invariant, meaning that $\bm{A}$, $\bm{B}$, $\bm{C}$ are constant through time, instead of depending on the provided context. 
Improved expressiveness was introduced with selective state space models and the Mamba-SSM architecture, which removes linear time invariance, boosts parallel efficiency, and uses selection to improve input context awareness~\cite{gu2024mambalineartimesequencemodeling}. 
We leverage this architecture to condition the normalizing flow on past and future observations. 
The state space model context encoder maps the normalized output embedding of the Mamba-SSM to the desired dimensionality used in conditioning the normalizing flow. 
A diagram adapted from \cite{gu2024mambalineartimesequencemodeling} is shown in Figure~\ref{fig:nf_cond_mamba}.
\rev{We use a single file Mamba implementation from~\cite{ip2024mamba2minimal}.}

% =====================================================
\subsubsection{Optimal Transport-inspired Kinetic Term}\label{sec:kinetic}
% =====================================================

Optimal transport provides a mathematical framework for finding the most efficient way to transform one probability distribution into another~\cite{villani2021topics}.
In this context, a normalizing flow can be understood as learning a map (realized through a composition of diffeomorphisms) $\bm{f}\colon \mathbb{R}^{d}\to\mathbb{R}^{d}$ that transforms samples $\bm{z}$ from the base distribution to samples $\bm{x}$ from the target density.
In other words, a NF defines a \emph{path} between the base and target probability distribution. Paths associated with minimal Kinetic Energy (KE) lead to Optimal Transport (OT) interpolants, characterized by minimal Wasserstein distance with respect to the base and target densities.
Additionally, since a NF architecture is typically overparametrized, regularization in terms of KE can mitigate the effects of overparameterization. 

Therefore, inspired by recent research in optimal transport~\cite{villani2021topics}, we add a \emph{kinetic energy} term to the loss function~\cite{shaul2023kineticoptimalprobabilitypaths,HUANG2023112155}. 
Formally, we augment the standard maximum likelihood objective used in training normalizing flows with a kinetic regularization term and a layer-wise prior matching penalty
\begin{equation}\label{equ:kinetic}
\begin{split}
\inf_{\bm{\theta}}\, \mathcal{L}(\bm{\theta}) 
= 
&\lambda_{1} \,
\mathbb{E}_{\bm{x} \sim p_{\bm{X}}(\bm{x})}
\!\left[-\log p_{\bm{X}}(\bm{x}; \bm{\theta})\right] + 
\lambda_{2} \, \frac{1}{L-1}
\sum_{\ell=1}^{L-1}
\mathbb{E}_{\bm{x} \sim p_{\bm{X}}(\bm{x})}
\!\left[\big\| f_{\ell+1}(\bm{x}) - f_{\ell}(\bm{x}) \big\|_2 \right]\\[4pt]
+
&\lambda_{3} \, \frac{1}{L-1}
\sum_{\ell=1}^{L-1}
\mathbb{E}_{\bm{x} \sim p_{\bm{X}}(\bm{x})}
\!\left[-\log p_{\bm{Z}}(f_{\ell}(\bm{x}))\right]
\end{split}
\end{equation}
where $\bm{f}_{l},\,l=1,\dots,L$ are the members of the NF collection of transformations and $\lambda_1$, $\lambda_2$, and $\lambda_3$ are regularization penalties. 
In equation~\eqref{equ:kinetic}, the first term corresponds to the negative log-likelihood of the entire composition of transformations, whereas the second term acts as a kinetic regularizer, encouraging smooth transitions between consecutive flow layers. Finally, the third term  maximizes the likelihood for intermediate layer outputs \rev{under the base distribution}.
For a preliminary assessment of the effects produced by KE regularization, we train an unconditional NF to estimate the well-known \emph{double moon} distribution~\cite{sklearn11} with and without considering the KE term and report the results in Figure~\ref{fig:nf_ke_prelim}. As expected, KE regularization leads to smoother probability paths, minimizing sample movement between two successive normalizing flow layers.
\rev{
 \begin{figure}[!ht]
    \centering
    % Left Subfigure: Without Kinetic Loss
    \begin{subfigure}[b]{0.46\textwidth}
        \centering
        \includegraphics[width=\linewidth]{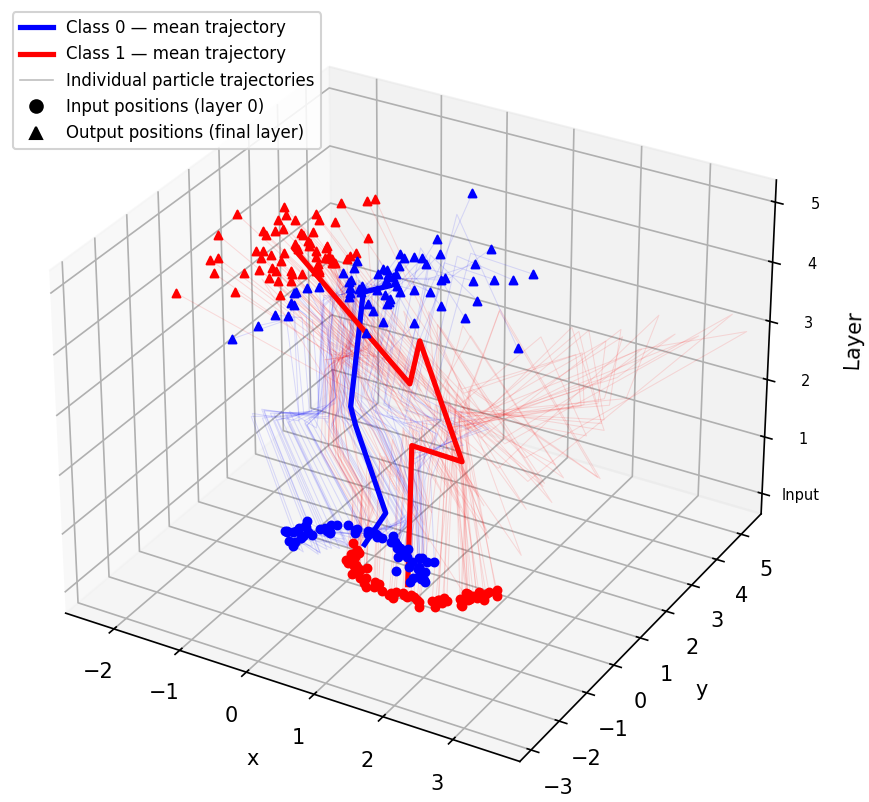} % Ensure you have a cropped version for the top/left
        \caption{Trained without kinetic loss.}
        \label{fig:nf_no_ke}
    \end{subfigure}
    $\quad$ % Pushes the subfigures to the absolute left and right, adding flexible space between
    % Right Subfigure: With Kinetic Loss
    \begin{subfigure}[b]{0.46\textwidth}
        \centering
        \includegraphics[width=\linewidth]{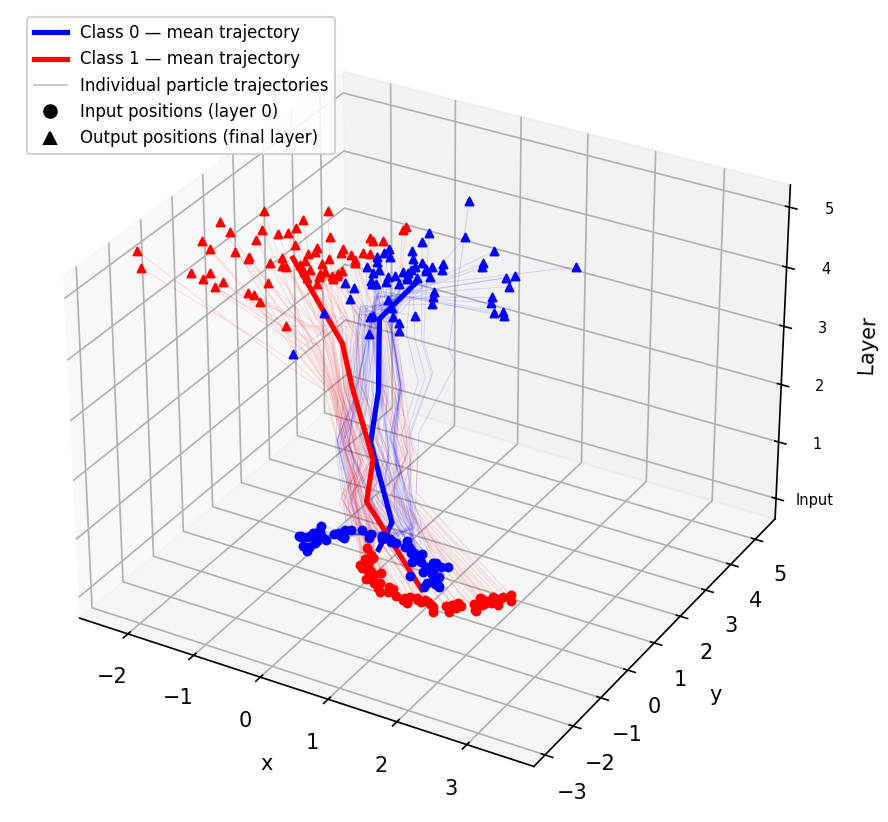} % Ensure you have a cropped version for the bottom/right
        \caption{Trained with kinetic loss.}
        \label{fig:nf_with_ke}
    \end{subfigure}

    \caption{\rev{Unconditional normalizing flow transform trained without (a) and with (b) kinetic loss term. Each plot is obtained by extracting the distribution of the output of each layer in the forward mapping $\boldsymbol{f(\cdot, \theta)}$ from the target density $\boldsymbol{p_X(X)}$ to the base distribution $\boldsymbol{p_Z(Z)}$. Note the effect of input parameter swap due to a permutation being performed between successive layers.}}
    \label{fig:nf_ke_prelim}
\end{figure}
}

 % ===================================================================
\subsubsection{Evaluating Distances using Kullback–Leibler Divergence}\label{sec:kl}
% ===================================================================

To evaluate the performance of conditional NF in forward and backward state estimation, we perform repeated calculations of the Kullback–Leibler (KL) divergence between two distributions that are not necessarily evaluated over the same samples.
While, in principle, one can use density estimation on both distributions and then evaluate them at a common set of samples, this will incur a significant computational cost. 
We instead use the following estimate suggested in~\cite{perez2008kullback, delecki2023deepnormalizingflowsstate}. Given $\bm{z}\in\mathbb{R}^{d}$, $n$ samples from $\widehat{p}(\bm{z})$, and $m$ samples from $p(\bm{z})$, one has
\begin{equation}
D_{KL}(\widehat{p}(\bm{z})\,  ||\,  p(\bm{z})) \approx \frac{d}{n}\, \sum_{i=1}^n log \frac{r_k(\bm{z}^{(i)})}{s_k(\bm{z}^{(i)})} + log\left(\frac{m}{n-1}\right),
\end{equation}
where $r_k(\bm{z}^{(i)})$ and $s_k(\bm{z}^{(i)})$ are the Euclidean distance to the $k$-th nearest neighbor of $\bm{z}^{(i)}$ in the samples from $\widehat{p}(\bm{z})$ and $p(\bm{z})$, respectively.

% Justification for KL_Divergence
\rev{
We adopt this evaluation method because, for the autonomous vehicle dataset with random switching, the true state distribution is not available in closed form: the velocity and angular acceleration are themselves random, introducing uncertainty that propagates through the vehicle's motion. Consequently, when comparing model predictions with true observations, neither is associated with an explicitly known distribution. This method therefore allows us to compare two distributions directly from their samples, bypassing density estimation.
}

% ======================
\subsubsection{Training}\label{sec:nf_training}
% ======================

\rev{For the autonomous vehicle dataset, w}e trained a normalizing flow with 10 layers, each containing an autoregressive transformation with 2 inputs, 4 hidden features, 4 context variables, with no batch normalization applied to the outputs. Also, a 2-layer MLP is used to encode the transformer or mamba-based embedding onto the parameters of the base distribution.
Such embedding is created using five sequential observations in time, adding Gaussian noise with $\sigma = 1.0$.
We perform 10,000 training iterations, using Adam, a batch size of 2,048 and a constant learning rate equal to $0.001$. 
\rev{For only the sparse autonomous vehicle dynamics with random switch dataset, we instead train a normalizing flow with $4$ layers, 5,000 training iterations, and a batch size of 128 for better generalization.
We train each conditional normalizing flow using maximum likelihood estimation.
}
Figure~\ref{fig:nf_loss_train} shows a representative training loss profile for two conditioning operators (i.e., transformer- and mamba-based) as well as combining transformer conditioning plus KE loss augmentation for normalizing flow. \rev{Figure~\ref{fig:nf_loss_val} shows validation losses across training epochs for transformer conditioning and transformer conditioning plus KE loss augmentation. The inclusion of KE regularization loss term results in a lower minimum, overall mean, and last ten epoch mean validation loss compared to transformer conditioning alone.

% SIR dataset training hyperparams
For the epidemiological SIR dataset, we trained a normalizing flow with the same hyperparameters listed previously except the autoregressive transformation has 3 inputs, 6 hidden features, 6 context variables, and the embedding is created using five sequential observations in time, adding Gaussian noise with $\sigma = 0.01$.
}
\rev{
\begin{figure}[!ht]
    \centering

    % ---- Left Subfigure: Training Loss ----
    \begin{subfigure}{0.48\linewidth}
        \centering
        \includegraphics[width=\linewidth]{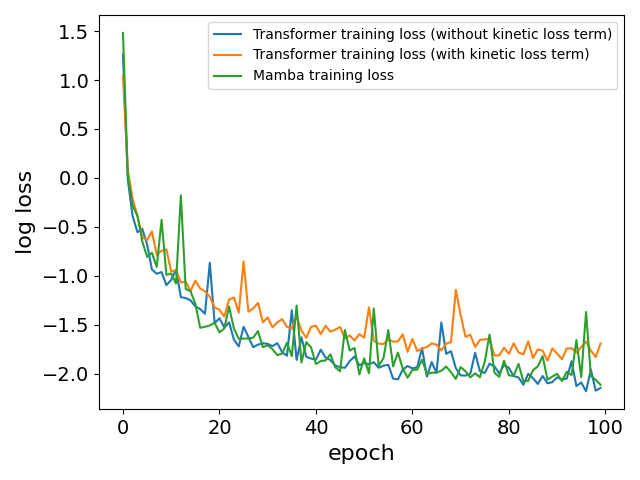}
        \caption{Training loss for different normalizing flow conditioning operators.}
        \label{fig:nf_loss_train}
    \end{subfigure}
    \hfill % Adds horizontal space to push subfigures to the margins
    % ---- Right Subfigure: Validation Loss ----
    \begin{subfigure}{0.48\linewidth}
        \centering
        \includegraphics[width=\linewidth]{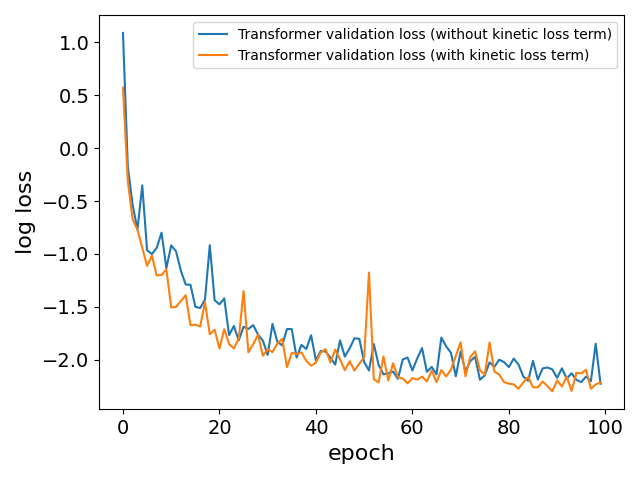}
        \caption{Validation loss for transformer conditioning operator with and without KE loss term.}
        \label{fig:nf_loss_val}
    \end{subfigure}

    \caption{Comparison of training and validation loss across epochs.}
    \label{fig:nf_loss_combined}
\end{figure}
 }

% ==================================
\section{Results}\label{sec:results}
% ==================================

% =========================================================
\subsection{Autonomous Vehicle Dynamics with Random Switch}
% =========================================================

% ===================================================
\rev{\subsubsection{Comparison with Baseline Methods}
% ===================================================

After training the conditional NFs, forward state estimation is performed by providing $R=5$
noisy observations of past vehicle locations as context, and estimating the conditional density of the next observation in the sequence.
Formally, given an initial observation at discrete time $n$, we estimate the density 
\begin{equation}
    \pi_{n+5} = p(\bm{x}_{n+5} | \bm{o}_{n}, \bm{o}_{n+1},\dots, \bm{o}_{n+4}).
\end{equation}
For the SMC algorithm (see Section~\ref{sec:pf}) we estimate $\pi_{n+5}$ by sampling the prior $p(\bm{o}_n)$, iterating through four steps of the algorithm, and propagating the particles one step forward with no observation update.
The performance of both SMC and NF is quantified by computing the mean KL divergence between 1,000 samples drawn from the true distribution of $\bm{x}_{n+5}$ and the predicted density (or SMC particles) at three locations along the trajectory, i.e., before, at and after the bifurcation. 
%
%In this comparison, the conditional NF is trained on the autonomous vehicle with random switching dataset with fixed time steps, and the SMC algorithm filters a trajectory from the same dataset.
%
Results for a dataset with constant time step are shown in Table~\ref{tab:res_vehicle_baseline}.
In all cases, the transformer-conditioned NF achieves a lower KL divergence than the SMC baseline, demonstrating its superior ability to approximate multi-modal filtering distributions.
Additionally, Figure~\ref{fig:nf_smc_comparison} shows samples for the forecast distribution generated by the two approaches. 
While NF is able to capture a multi-modal distribution at the bifurcation, SMC particles appear to concentrate on a single mode. 

% Other differences with respect to SMC
We also highlight two important differences between the proposed approach and SMC. First, SMC requires explicit knowledge of the state update equation; consequently, for stochastic systems, a characterization of the random trajectories must be provided to the algorithm at every step. 
Second, the proposed approach supports a \emph{dropout} mechanism that effectively mimics a variable time step, whereas SMC typically requires observations acquired at constant time steps. 

\begin{table}[ht]
    \centering
    \renewcommand{\arraystretch}{1.5} % Increases vertical padding
    \setlength{\tabcolsep}{10pt}      % Increases horizontal padding
    \begin{tabular}{|c|l|c|c|}
        \hline
        \multicolumn{2}{|c|}{} & \multicolumn{2}{c|}{\textbf{Model}} \\
        \cline{3-4}
        \multicolumn{2}{|c|}{} & \textbf{Conditional NF} & \textbf{Sequential MC} \\
        \hline
        \multirow{3}{*}{\rotatebox{90}{\textbf{Sample}}} 
        & \textbf{Before-bifurcation (fw)} & \textbf{0.843} & 7.554 \\
        \cline{2-4}
        & \textbf{At-bifurcation (fw)} & \textbf{3.268} & 8.856 \\
        \cline{2-4}
        & \textbf{After-bifurcation (fw)} & \textbf{0.777} & 9.069 \\
        \hline
    \end{tabular}
    \caption{Mean KL divergence results comparing conditional normalizing flow to the SMC baseline.}\label{tab:res_vehicle_baseline}
\end{table}

 \begin{figure}[!ht]
    \centering
    % Left Subfigure: Without Kinetic Loss
    \begin{subfigure}[b]{0.46\textwidth}
        \centering
        \includegraphics[width=\linewidth]{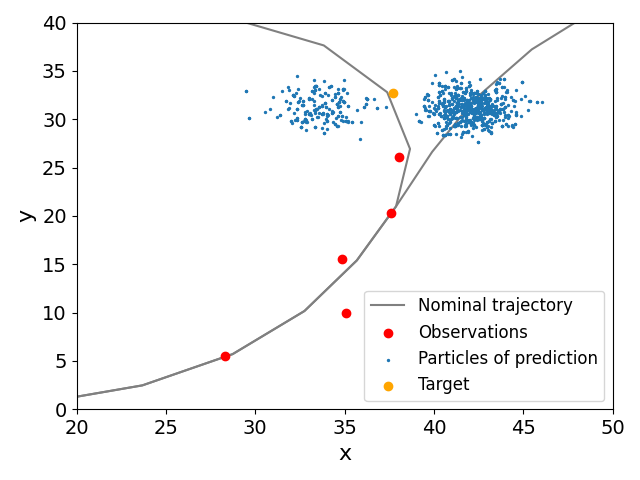} % Ensure you have a cropped version for the top/left
        \caption{Particles from SMC propagated one step forward.}
        \label{fig:pf_particles}
    \end{subfigure}
    $\quad$ % Pushes the subfigures to the absolute left and right, adding flexible space between
    % Right Subfigure: With Kinetic Loss
    \begin{subfigure}[b]{0.46\textwidth}
        \centering
        \includegraphics[width=\linewidth]{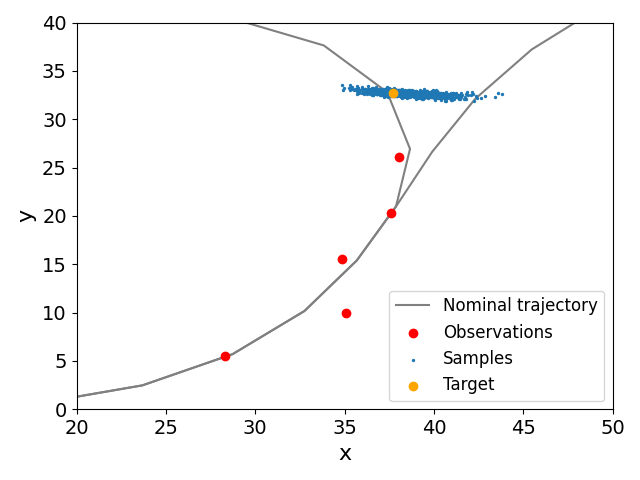} % Ensure you have a cropped version for the bottom/right
        \caption{Samples from estimated distribution using NF.}
        \label{fig:nf_samples}
    \end{subfigure}

    \caption{\rev{Comparison of samples estimated using (a) SMC and (b) NF, respectively. Both approaches were trained on the autonomous vehicle dataset using a constant time step.}}
    \label{fig:nf_smc_comparison}
\end{figure}

}

% ==========================================================================
\rev{\subsubsection{Autonomous Vehicle Dynamics with Random Switch Results}}
% ==========================================================================

After training the conditional NFs \rev{using the full dataset with variable time steps, show in Figure~\ref{fig:autonomous_data_b}}, forward state estimation is performed by providing $R=5$ previous noisy observations for the vehicle location as context, and estimating the conditional density for the next observation in sequence. 
In other words, assuming an initial observation at discrete time $n$, we estimate the density $\rev{\pi_{n+5}}=p(\bm{x}_{n+5} | \bm{o}_{n}, \bm{o}_{n+1},\dots, \bm{o}_{n+4})$.
Similarly, for backward state estimation we provide $R=5$ future noisy vehicle positions and estimate the density for the previous observation in the sequence, or
\begin{equation}
\rev{\pi_{n-5}} = p(\bm{x}_{n-5} | \bm{o}_{n}, \bm{o}_{n-1},\dots, \bm{o}_{n-4})
\end{equation}

\begin{figure}[!ht]
     \centering
     \includegraphics[width=1\linewidth]{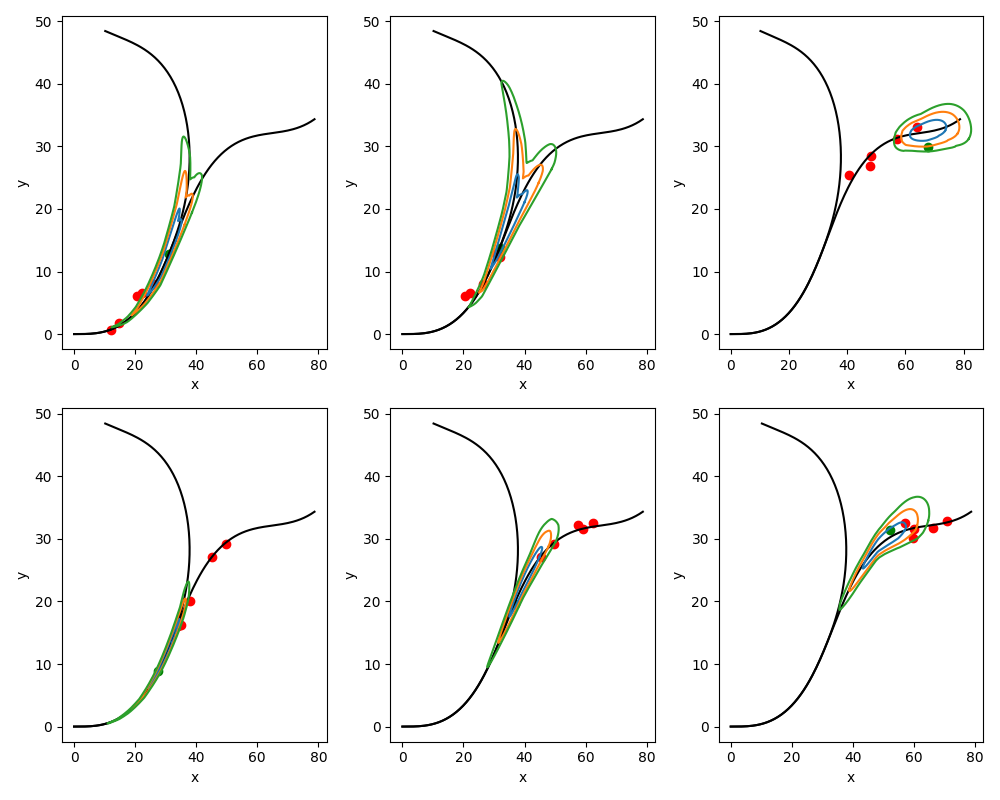}
     \caption{Confidence regions corresponding to 1, 2, and 3 base distribution standard deviations for forward (top) and backward (bottom) state estimation, for NF with transformer-based conditioning. Predicted density refers to three locations along the trajectory (left to right).}
     \label{fig:res_traj_transformer}
 \end{figure}

\begin{figure}[!ht]
     \centering
     \includegraphics[width=1\linewidth]{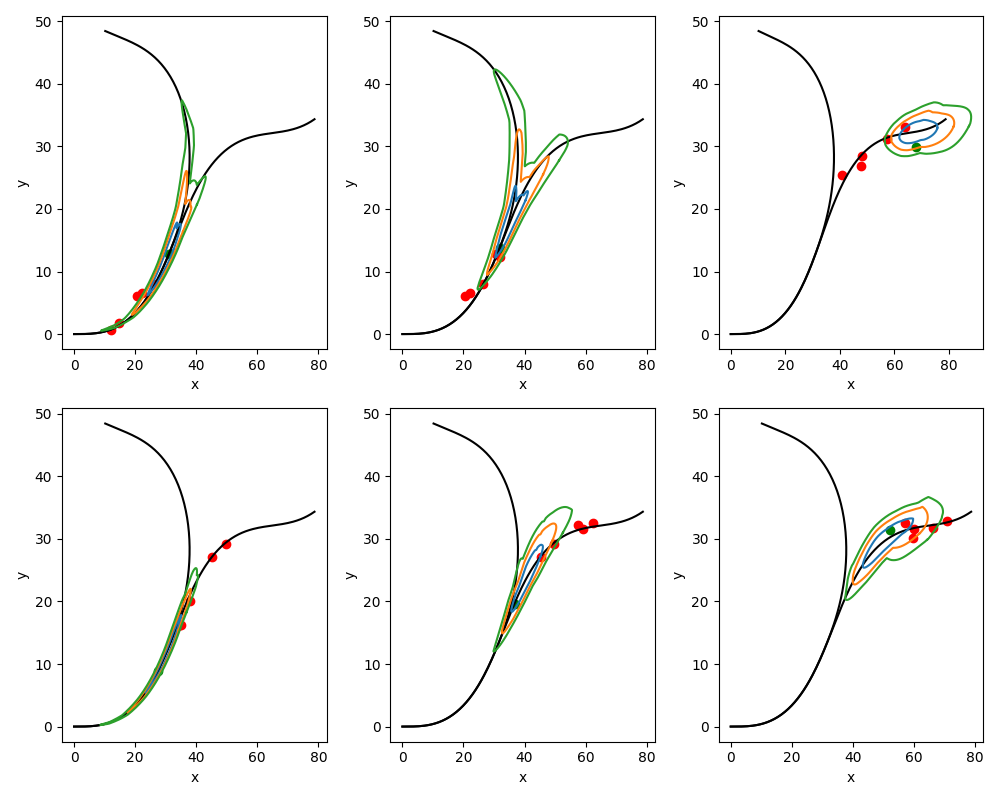}
     \caption{Confidence regions corresponding to 1, 2, and 3 base distribution standard deviations for forward (top) and backward (bottom) state estimation, for NF with transformer-based conditioning, plus KE loss term. Predicted density refers to three locations along the trajectory (left to right).}
     \label{fig:res_traj_transformer_ke}
 \end{figure}

\begin{figure}[!ht]
     \centering
     \includegraphics[width=1\linewidth]{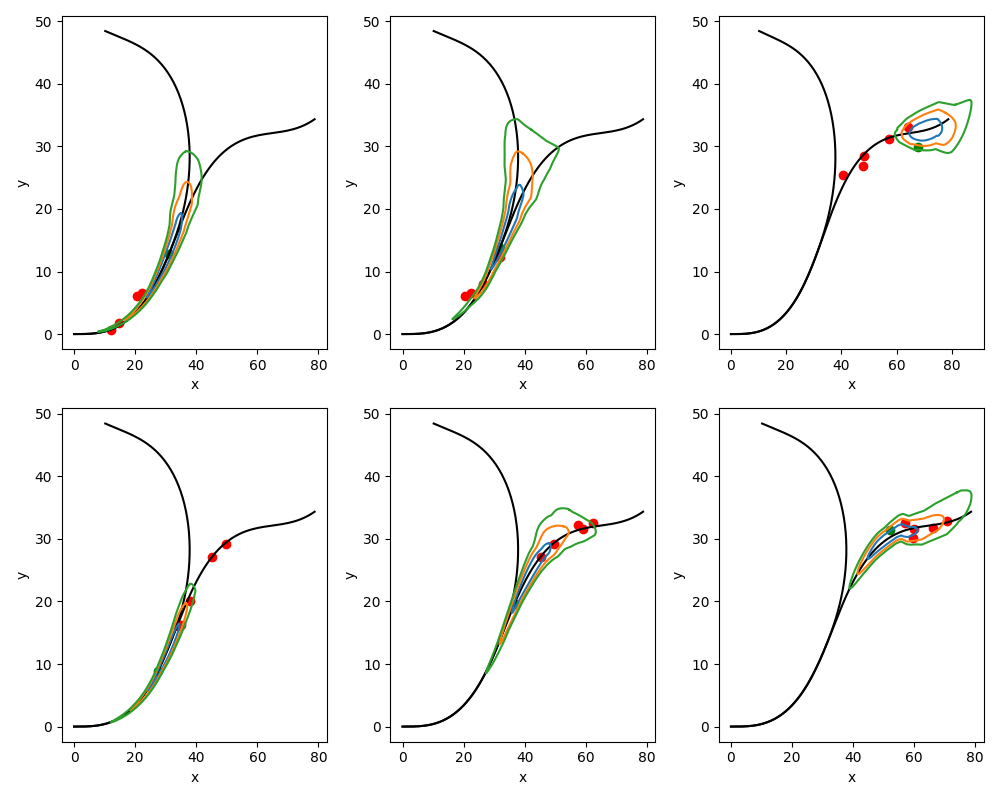}
     \caption{Confidence regions corresponding to 1, 2, and 3 base distribution standard deviations for forward (top) and backward (bottom) state estimation, for NF with Mamba-SSM-based conditioning. Predicted density refers to three locations along the trajectory (left to right).}
     \label{fig:res_traj_mamba}
 \end{figure}

We also evaluate the effectiveness of forward and backward state estimation for contexts that are provided at three different locations along the bimodal trajectory. Specifically, we consider contexts that include locations before the bifurcation, at the bifurcation, and after the bifurcation as shown in Figure~\ref{fig:res_traj_transformer}, \ref{fig:res_traj_transformer_ke}, and \ref{fig:res_traj_mamba}, respectively.  

In addition, we calculate the KL divergence between 1,000 samples from the estimated conditional density and the true distributions of $\bm{x}_{n+5}$ and $\bm{x}_{n-5}$ at the three locations along the trajectory. 
Aggregated results are shown in Table~\ref{tab:res_vehicle}. As previously shown in Figure~\ref{fig:nf_ke_prelim}, the addition of the kinetic loss term results in smoother transformations from the base to the target distribution, leading to improved accuracy. 
Such configuration better captures the true conditional density without exhibiting bias towards a certain mean trajectory in the bifurcation, thus leading to a lower average KL-divergence aggregated over many noisy samples \rev{at the bifurcation for the NF with transformer-based conditioning with KE loss}.
\rev{Since the data does follow a mean trajectory before and after the bifurcation, it explains why the NF with transformer-based conditioning, without KE loss, and the Mamba-SSM conditioning operator exhibit a lower average KL-divergence aggregated over many noisy samples.}

\renewcommand{\arraystretch}{1.5} 

\begin{table}
    \centering
    \begin{tabular}{|c|c|c|c|c|}
    \hline
    \multicolumn{2}{|c|}{} &\multicolumn{3}{c|}{\bf Model} \\
     \cline{3-5}
     \multicolumn{2}{|c|}{}  & {\bf Transformer} & {\bf Transformer + KE} & {\bf Mamba-SSM} \\\hline
      \multirow{6}{*}{\makecell{\rotatebox{90}{\bf Sample}}}  & {\bf Before-bifurcation (fw)} & \( 5.062\) &  { 4.989}&  \(\bf 4.988\) \\\cline{2-5}    
        & {\bf Before-bifurcation (bw)} & \(4.955\) &  { 4.893} &  \(\bf4.621\) \\\cline{2-5}
       & {\bf At-bifurcation (fw)} & \(4.503\) &  \(\bf 4.484\) &  \( 4.572\) \\\cline{2-5}
       & {\bf At-bifurcation (bw)} & \(4.844\) &  \({ 4.827}\) &  \(\bf 4.681\) \\\cline{2-5}
       & {\bf After-bifurcation (fw)} & {\bf 3.719} &  { 3.791} &  \(3.753\) \\\cline{2-5}
       & {\bf After-bifurcation (bw)} & \(4.502\) &  \({ 4.463}\) &  \(\bf 4.157\) \\
       \hline  
      \end{tabular}
    \caption{Mean KL divergence between 1,000 samples from estimated conditional density and true distribution when predicting next and previous states at three different locations along the vehicle trajectory.}\label{tab:res_vehicle}
\end{table}

% =======================================
\subsubsection{Training with Sparse Data}
% =======================================

\rev{To show that our method may not necessarily rely on millions of observations, we train a conditional NF with transformer conditioning operator on the sparse autonomous vehicle with random switching dataset, shown in Figure~\ref{fig:autonomous_data_c}. We consider contexts that include locations before the bifurcation, at the bifurcation, and after the bifurcation. Figure~\ref{fig:res_traj_transformer_sparse} shows the models ability to capture the conditional distribution even when trained on sparse data.

\begin{figure}[!ht]
     \centering
     \includegraphics[width=1\linewidth]{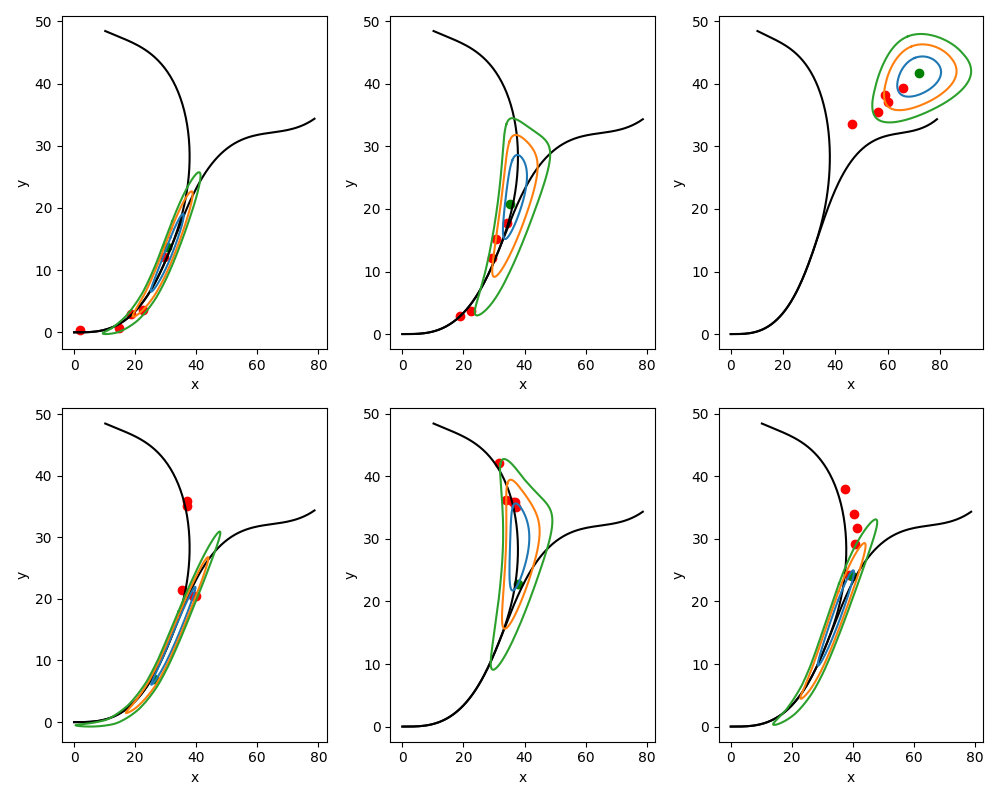}
     \caption{\rev{Confidence regions corresponding to 1, 2, and 3 base distribution standard deviations for forward (top) and backward (bottom) state estimation, for NF with transformer-based conditioning trained on sparse dataset. Predicted density refers to three locations along the trajectory (left to right).}}
     \label{fig:res_traj_transformer_sparse}
 \end{figure}
}

% ====================
\subsection{SIR Model}\label{sec:SIR_res}
% ====================

While the advantage of performing joint forward and backward conditional state estimation may not be readily apparent in the context of autonomous vehicles, it is certainly evident with epidemiological data.
For applications in epidemiology, making past and future estimates for the number of susceptible, infected or recovered individuals from a snapshot of uncertain observations in time can lead to an improved understanding of disease progression dynamics.
Also, estimates of the progression of the underlying system parameters are essential to inform public health policies.

In Figure~\ref{fig:transformer_forward_single} we visualize $1\sigma$ uncertainty regions for forward state density estimation resulting from training using a single SIR model solution (one set of $\beta$, $\gamma$ and initial conditions). \rev{The illustration in Figure~\ref{fig:transformer_forward_single} considers only the transformer-based conditioning operator, but we aggregate results from all three proposed architectures.}
Table~\ref{tab:res_sir_single_run_l2_split} shows the mean KL divergence between 1,000 samples from estimated conditional densities and the true underlying distributions at randomly chosen points along the SIR model trajectory. 
%The results demonstrate that all conditional flows achieve a low KL divergence when applied to the SIR model compared to the multimodal autonomous driving dataset.
Among the conditioning operators, Mamba-SSM achieves the lowest KL-divergence in all compartments for both forward and backwards state estimation. We attribute this performance to the temporal dependencies and unimodal nature of the SIR trajectories. Mamba's selective state-space architecture is particularity well-suited to capture a trajectory governed by ODEs because it maintains a latent representation of the system, thus explaining its strong performance compared to the transformer architectures which rely more on local context.

\rev{
\begin{figure}[!ht]
    \centering
    \includegraphics[width=0.75\linewidth]{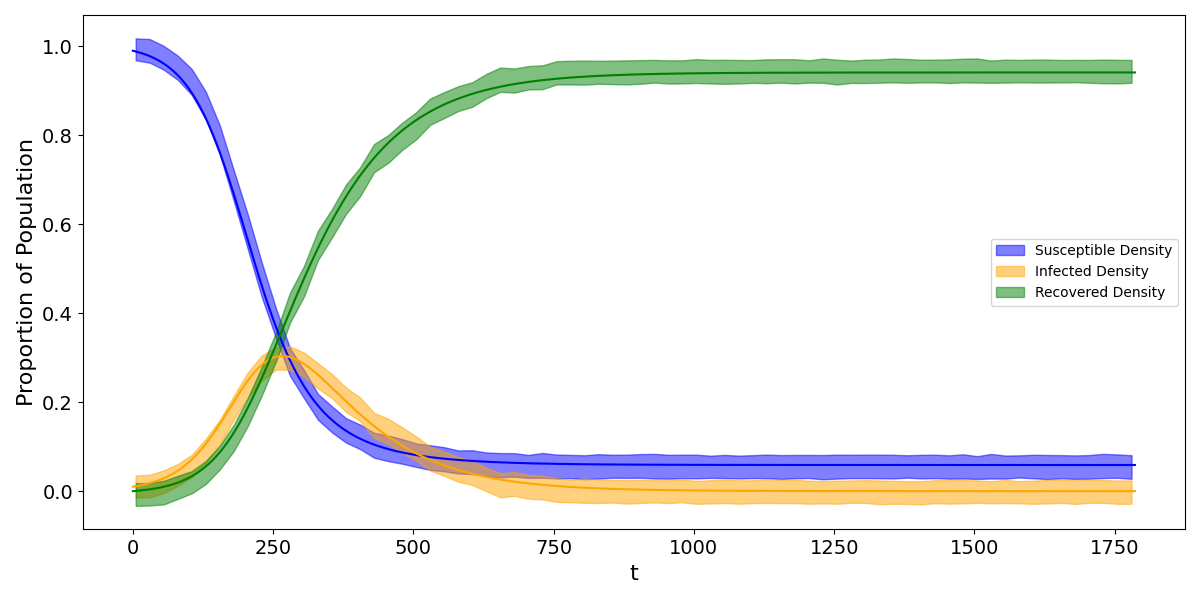} % Adjusted width for singular figure
    \caption{\(1\sigma\) error bounds for forward (fw) state predictions given a five-step observation context at various points in the SIR system using NF with transformer conditioning operator.}
    \label{fig:transformer_forward_single}
\end{figure}

}

\renewcommand{\arraystretch}{2.25}  % Increase vertical spacing

\begin{table}[h!]
    \centering
    \begin{tabular}{|c|c|ccc|ccc|ccc|}
    \hline
    \multicolumn{2}{|c|}{} & \multicolumn{9}{c|}{\bf Model} \\
    \cline{3-11}
    \multicolumn{2}{|c|}{} & \multicolumn{3}{c|}{\bf Transformer} & \multicolumn{3}{c|}{\bf Transformer + KE} & \multicolumn{3}{c|}{\bf Mamba-SSM} \\
    \cline{3-11}
    \multicolumn{2}{|c|}{} & {\bf S} & {\bf I} & {\bf R} & {\bf S} & {\bf I} & {\bf R} & {\bf S} & {\bf I} & {\bf R} \\
    \hline
    \multirow{2}{*}{\makecell{\rotatebox{90}{\bf Sample}}}
        & {\bf fw} & 2.245 & 2.211 & 2.304 & 2.243 & 2.375 & 2.467 & \bf{2.192} & \bf{2.209} & \bf{2.264} \\
    \cline{2-11}
        & {\bf bw} & 2.110 & 1.984 & 2.232 & 2.143 & 2.098 & 2.320 & \bf{2.053} & \bf{1.975} & \bf{2.157} \\
    \hline
    \end{tabular}
    \caption{Mean KL divergence between 1,000 samples from the estimated conditional density and true distribution for forward and backward state estimation at 100 randomly sampled locations along the noisy SIR trajectory. Errors are shown separately for each compartment \(S\), \(I\), and \(R\).}
    \label{tab:res_sir_single_run_l2_split}
\end{table}

% =================================================
\subsubsection{Training with Multiple Trajectories}
% =================================================

After assessing the ability of conditional NF to accurately infer states from a single SIR trajectory, we study the ability of the system to work with multiple trajectories using the dataset shown in Figure~\ref{fig:sir_multiple_traj_ds}. 
The data set shows noisy susceptible, infected, recovered populations as a proportion of the total population over time. 

\begin{figure}[!ht]
    \centering

    % ---- Row 1: Forward ----
    \begin{subfigure}[b]{0.48\linewidth}
        \centering
        \includegraphics[width=\linewidth]{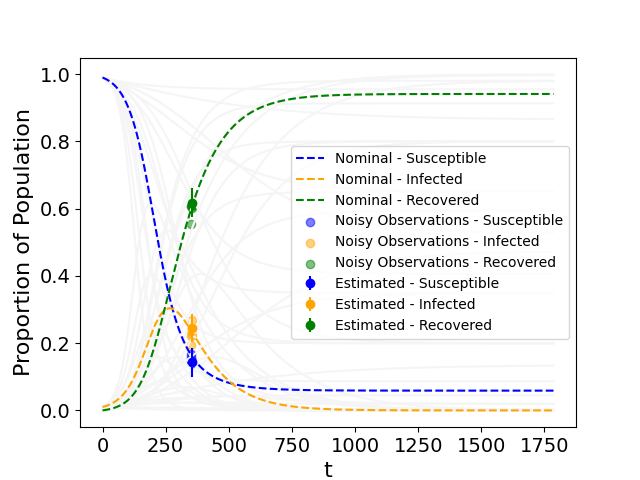}
        \caption{Forward state estimate (full view).}
        \label{fig:multi_traj_state_estimate_forward_full}
    \end{subfigure}
    \hfill
    \begin{subfigure}[b]{0.48\linewidth}
        \centering
        \includegraphics[width=\linewidth]{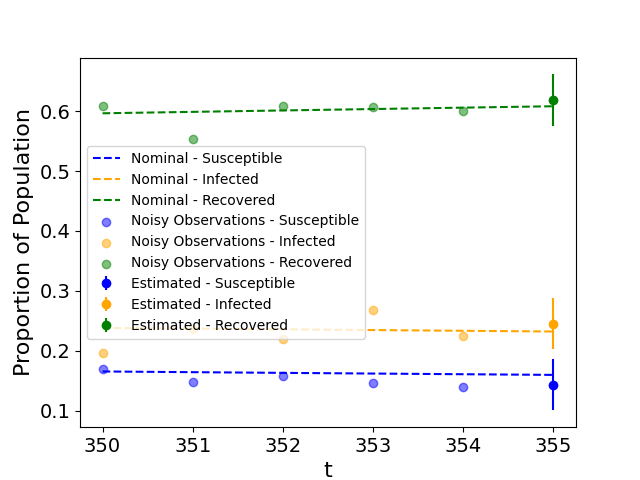}
        \caption{Forward state estimate (zoomed view).}
        \label{fig:multi_traj_state_estimate_forward_zoom}
    \end{subfigure}

    \vspace{0.8em}

    % ---- Row 2: Inverse ----
    \begin{subfigure}[b]{0.48\linewidth}
        \centering
        \includegraphics[width=\linewidth]{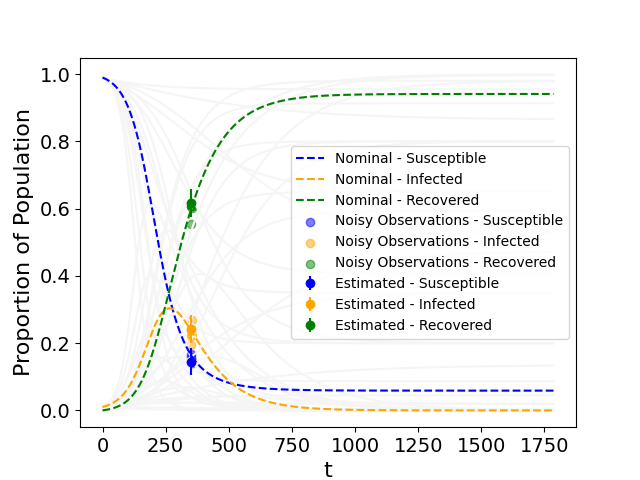}
        \caption{\rev{Backward} state estimate (full view).}
        \label{fig:multi_traj_state_estimate_inverse_full}
    \end{subfigure}
    \hfill
    \begin{subfigure}[b]{0.48\linewidth}
        \centering
        \includegraphics[width=\linewidth]{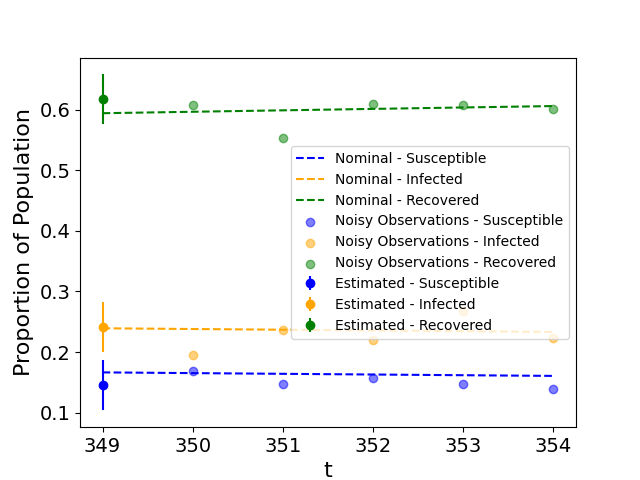}
        \caption{\rev{Backward} state estimate (zoomed view).}
        \label{fig:multi_traj_state_estimate_inverse_zoom}
    \end{subfigure}

    \caption{State estimates for normalizing flow conditioned on multiple trajectories using a transformer conditioning operator. 
    Each row shows the full view (left) and a zoomed-in region (right). 
    (a,b) Forward state estimates with noisy context. 
    (c,d) Backward state estimates with noisy context.}
    \label{fig:multi_traj_state_estimate_2x2}
\end{figure}

Figure~\ref{fig:multi_traj_state_estimate_2x2} shows the ability of NF in forward and backward state estimation given a noisy context from a single SIR model solution but trained  on multiple trajectories. 
We observe that the model is able to accurately predict the next state and correctly identify the singular underlying trajectory for the susceptible and recovered populations despite being trained on multiple trajectories. 
The infected portion of the population is predicted with relatively lower accuracy. This is likely due its reduced sensitivity to changes in the underlying system parameters, having a negative effect on the generalization abilities of the proposed approach.
Finally, Table~\ref{tab:res_sir_multi_traj_l2_split} shows the resulting mean KL divergence for this case. \rev{In most cases, the Mamba-SSM conditioning operator achieves the lowest KL-divergence, for reasons discussed in Section~\ref{sec:SIR_res}. However, in certain cases, transformer-based conditioning with KE loss outperforms, showing the ability of the kinetic loss term to estimate accurate conditional densities in complex systems.}  

% Old multiple trajectory plot
% \begin{figure}[!ht]
%      \centering
%      \includegraphics[width=0.8\textwidth]{fig_9.png}
%      \caption{Ensemble of noisy SIR model solutions for dataset with multiple trajectories.}
%      \label{fig:sir_multiple_traj_ds}
%  \end{figure}

\rev{
% New multiple trajectory plot
\begin{figure}[!ht]
    \centering
    \begin{subfigure}[b]{0.32\textwidth}
        \centering
        \includegraphics[width=\textwidth]{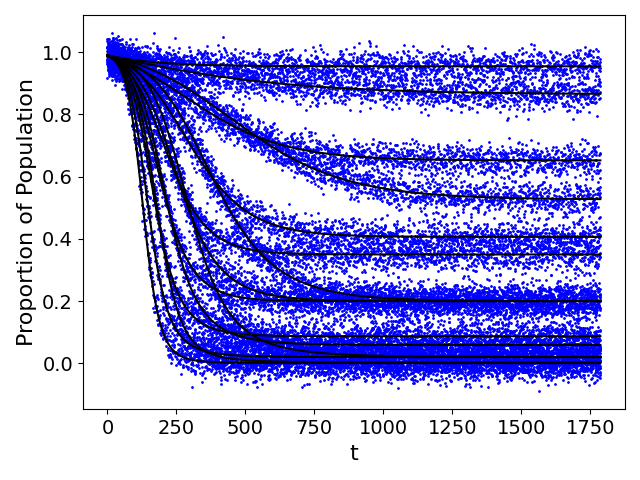}
        \caption{Susceptible}
        \label{fig:sir_multiple_traj_ds_S}
    \end{subfigure}
    \hfill
    \begin{subfigure}[b]{0.32\textwidth}
        \centering
        \includegraphics[trim={0pt -6pt 0pt 0pt}, clip,width=\textwidth]{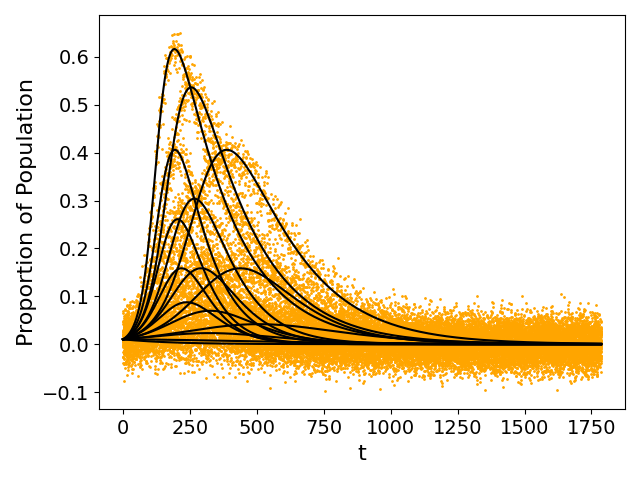}
        \caption{Infected}
        \label{fig:sir_multiple_traj_ds_I}
    \end{subfigure}
    \begin{subfigure}[b]{0.32\textwidth}
        \centering
        \includegraphics[trim={0pt -6pt 0pt 0pt}, clip,width=\textwidth]{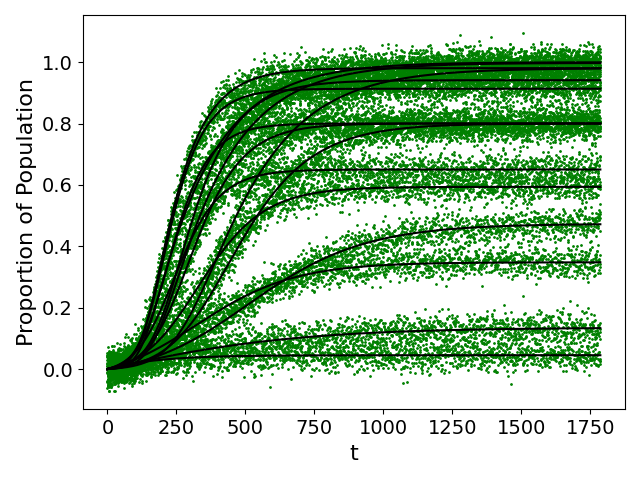}
        \caption{Removed}
        \label{fig:sir_multiple_traj_ds_R}
    \end{subfigure}
    \caption{Ensemble of noisy SIR model solutions for dataset with multiple trajectories separated by compartment.}
    \label{fig:sir_multiple_traj_ds}
\end{figure}
}
 
\begin{table}[h!]
    \centering
    \begin{tabular}{|c|c|ccc|ccc|ccc|}
    \hline
    \multicolumn{2}{|c|}{} & \multicolumn{9}{|c|}{\bf Model} \\
    \cline{3-11}
    \multicolumn{2}{|c|}{} & \multicolumn{3}{|c|}{\bf Transformer} & \multicolumn{3}{|c|}{\bf Transformer + KE} & \multicolumn{3}{|c|}{\bf Mamba-SSM} \\
    \cline{3-11}
    \multicolumn{2}{|c|}{} & {\bf S} & {\bf I} & {\bf R} & {\bf S} & {\bf I} & {\bf R} & {\bf S} & {\bf I} & {\bf R} \\
    \hline
    \multirow{2}{*}{\makecell{\rotatebox{90}{\bf Sample}}}
        & {\bf fw} & 2.847 & 2.456 & 2.642 & 2.816 & \bf{2.440} & 2.616 & \bf{2.729} & 2.478 & \bf{2.418} \\
    \cline{2-11}
        & {\bf bw} & 2.838 & 2.097 & 2.828 & \bf{2.634} & 2.138 & 2.765 & 2.765 & \bf{2.051} & \bf{2.662} \\
    \hline
    \end{tabular}
    \caption{Mean KL divergence between 1,000 samples from estimated conditional density and true distribution. Forward and backward state estimation is performed at 100 randomly sampled locations along the noisy SIR trajectory. The conditional NF predictor is trained on multiple SIR trajectories with various conditioning operators. Errors are shown separately for each compartment \(S\), \(I\), and \(R\).}
    \label{tab:res_sir_multi_traj_l2_split}
\end{table}

% ==============================
\subsubsection{COVID-19 Dataset}
% ==============================

\begin{figure}[!ht]
     \centering
     \includegraphics[width=0.75\linewidth]{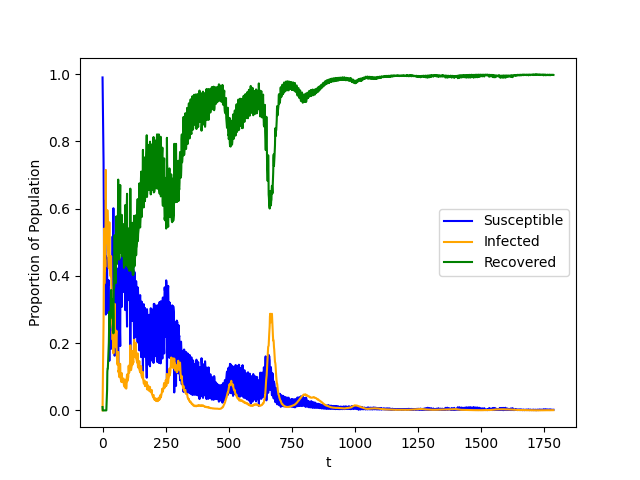}
     \caption{San Francisco COVID-19 SIR data.}
     \label{fig:sf_covid_data}
 \end{figure}

To further test the ability of NF to predict future trends with real data when trained from synthetic model solutions, we present a real-world state estimation task using COVID-19 data collected by the City and County of San Francisco Department of Public Health.
The dataset used for our analysis is available on DataSF, the Office of the Chief Data Officer~\cite{sfCOVID19Data}. 
Figure~\ref{fig:sf_covid_data} shows the generated SIR data from case, testing, and death reports, obtained using a similar data transformation process as in~\cite{AlQadi2022-jy}. 

First, we perform forward and backward state estimation using a context which consists of real COVID-19 SIR observations, using NF trained on synthetic SIR trajectories with parameters drawn as $\beta\sim\mathcal{U}[0.02, 0.04]$ and $\gamma\sim\mathcal{U}[0.005, 0.025]$ and initial conditions $S_0=0.99$, $I_0=\rev{0.00}$, and $R_0=\rev{0.01}$.

\begin{figure}[!ht]
    \centering
    % ---- Top Row ----
    \begin{minipage}[b]{0.48\linewidth}
        \centering
        \includegraphics[width=\linewidth]{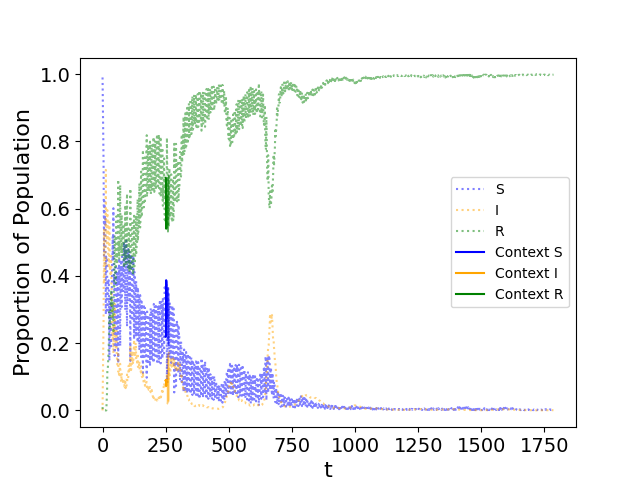}
        \caption*{\textbf{(a)} 7-day window \rev{(full view)}}
        \label{fig:sanfran_7day_fwd}
    \end{minipage}
    \hfill
    \begin{minipage}[b]{0.48\linewidth}
        \centering
        \includegraphics[width=\linewidth]{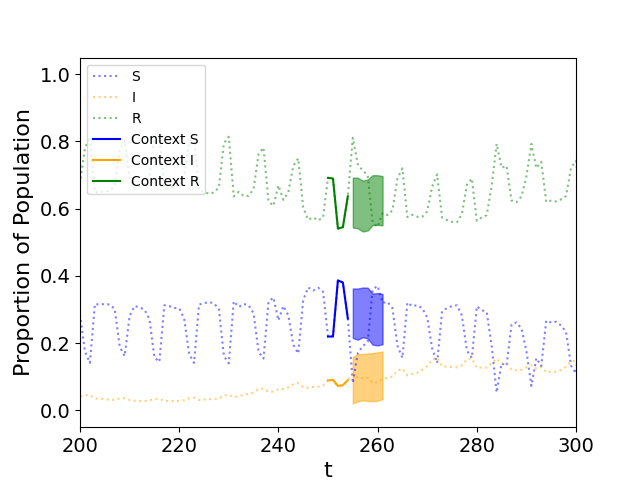}
        \caption*{\textbf{(b)} 7-day window (zoomed \rev{view})}
        \label{fig:sanfran_7day_fwd_zoomed}
    \end{minipage}

    \vspace{0.5em}

    % ---- Bottom Row ----
    \begin{minipage}[b]{0.48\linewidth}
        \centering
        \includegraphics[width=\linewidth]{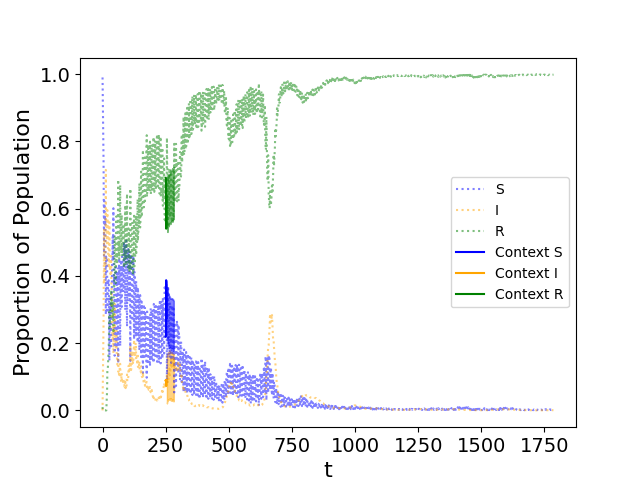}
        \caption*{\textbf{(c)} 28-day window \rev{(full view)}}
        \label{fig:sanfran_28day_fwd}
    \end{minipage}
    \hfill
    \begin{minipage}[b]{0.48\linewidth}
        \centering
        \includegraphics[width=\linewidth]{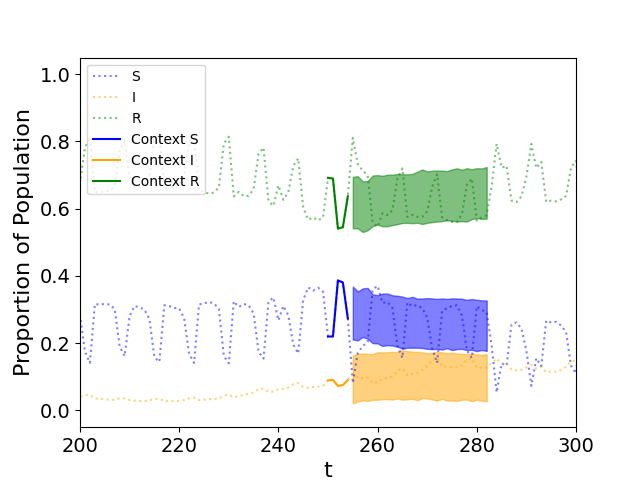}
        \caption*{\textbf{(d)} 28-day window (zoomed \rev{view})}
        \label{fig:sanfran_28day_fwd_zoomed}
    \end{minipage}

    \caption{Forward rollout predictions of San Francisco COVID-19 data with \(2\sigma\) error using a normalizing flow with transformer conditioning operator trained on multiple trajectories. Each row shows forward state estimation with different rolling window sizes: (a,b) 7-day and (c,d) 28-day.}
    \label{fig:combined_SANFRANCOVID19_rollouts_transformer_fwd}
\end{figure}

\begin{figure}[!ht]
    \centering
    % ---- Top Row ----
    \begin{minipage}[b]{0.48\linewidth}
        \centering
        \includegraphics[width=\linewidth]{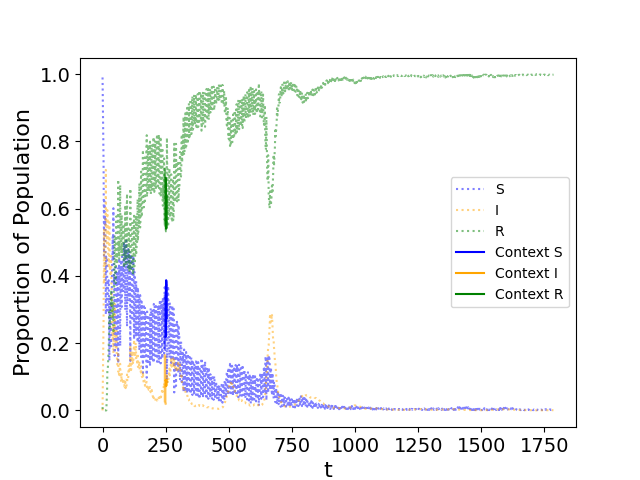}
        \caption*{\textbf{(a)} 7-day window \rev{(full view)}}
        \label{fig:sanfran_7day_inv}
    \end{minipage}
    \hfill
    \begin{minipage}[b]{0.48\linewidth}
        \centering
        \includegraphics[width=\linewidth]{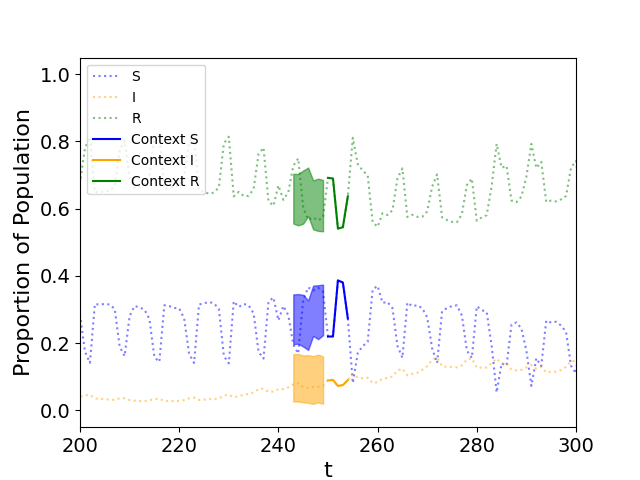}
        \caption*{\textbf{(b)} 7-day window (zoomed \rev{view})}
        \label{fig:sanfran_7day_inv_zoomed}
    \end{minipage}

    \vspace{0.5em}

    % ---- Bottom Row ----
    \begin{minipage}[b]{0.48\linewidth}
        \centering
        \includegraphics[width=\linewidth]{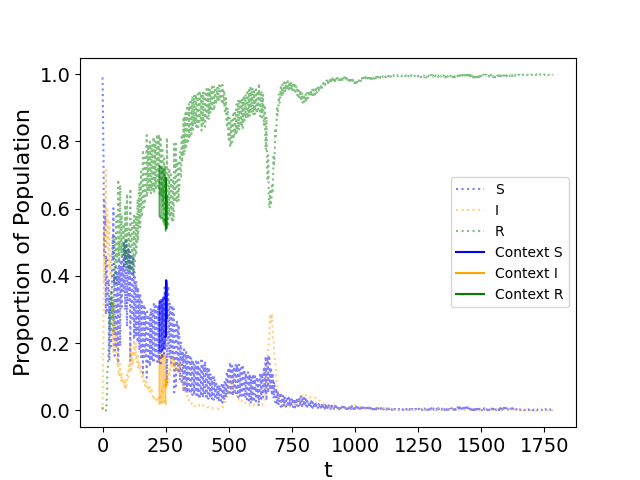}
        \caption*{\textbf{(c)} 28-day window \rev{(full view)}}
        \label{fig:sanfran_28day_inv}
    \end{minipage}
    \hfill
    \begin{minipage}[b]{0.48\linewidth}
        \centering
        \includegraphics[width=\linewidth]{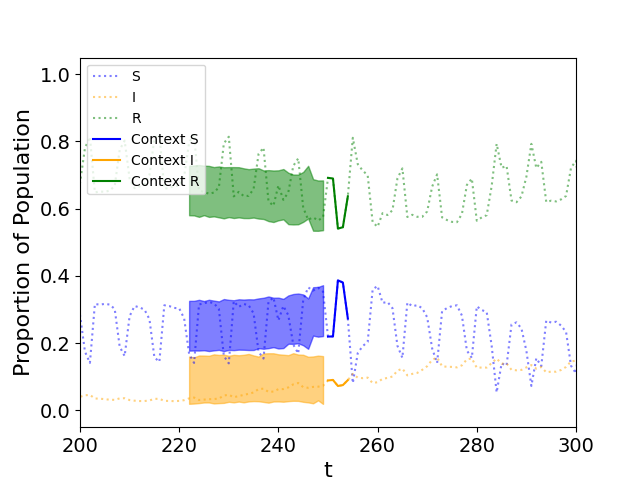}
        \caption*{\textbf{(d)} 28-day window (zoomed \rev{view})}
        \label{fig:sanfran_28day_inv_zoomed}
    \end{minipage}

    \caption{Backward rollout predictions of San Francisco COVID-19 data with \(2\sigma\) error using a normalizing flow with transformer conditioning operator trained on multiple trajectories. Each row shows backward state estimation with different rolling window sizes: (a,b) 7-day and (c,d) 28-day.}
    \label{fig:combined_SANFRANCOVID19_rollouts_transformer_inv}
\end{figure}

We then perform a forward and backward rollout exercise\rev{, as discussed in Section~\ref{sec:cond_embedding},} to test the ability of NF to predict future or past epidemiological trends. The results are reported in Figure~\ref{fig:combined_SANFRANCOVID19_rollouts_transformer_fwd} and Figure~\ref{fig:combined_SANFRANCOVID19_rollouts_transformer_inv} with $2\sigma$ error. The results in the figures show an ability of NF to predict future and past disease trends, particularly based on entirely synthetic knowledge. The mean negative log likelihood of the \rev{estimated conditional density given the true state} for the San Francisco COVID-19 test case is summarized in Table~\ref{tab:res_sir_COVID19_nll}.

\rev{
\begin{table}[h!]
    \centering
    \begin{tabular}{|c|c|c|c|c|}
    \hline
    \multicolumn{2}{|c|}{} & \multicolumn{3}{c|}{\bf Model} \\
    \cline{3-5}
    \multicolumn{2}{|c|}{} & {\bf Transformer} & {\bf Transformer + KE} & {\bf Mamba-SSM} \\
    \hline
    \multirow{2}{*}{\rotatebox{90}{\bf Sample}}
        & {\bf fw} & \bf{-2.468} & -2.398 & -1.795 \\
    \cline{2-5}
        & {\bf bw} & \bf{-2.231} & -2.092 & -1.427 \\
    \hline
    \end{tabular}
    \caption{Mean negative log likelihood of estimated conditional density given true observation. Forward and backward state estimation is performed at 100 randomly sampled locations along the San Francisco COVID-19 trajectory. The conditional NF predictor is trained on multiple SIR trajectories with various conditioning operators. Errors are aggregated across all compartments.}
    \label{tab:res_sir_COVID19_nll}
\end{table}
}

% ==================================================
\subsubsection{Predicting Disease System Parameters}
% ==================================================

\begin{figure}[!ht]
    \centering
    \begin{minipage}[b]{0.75\linewidth}
        \centering
        \includegraphics[width=\linewidth]{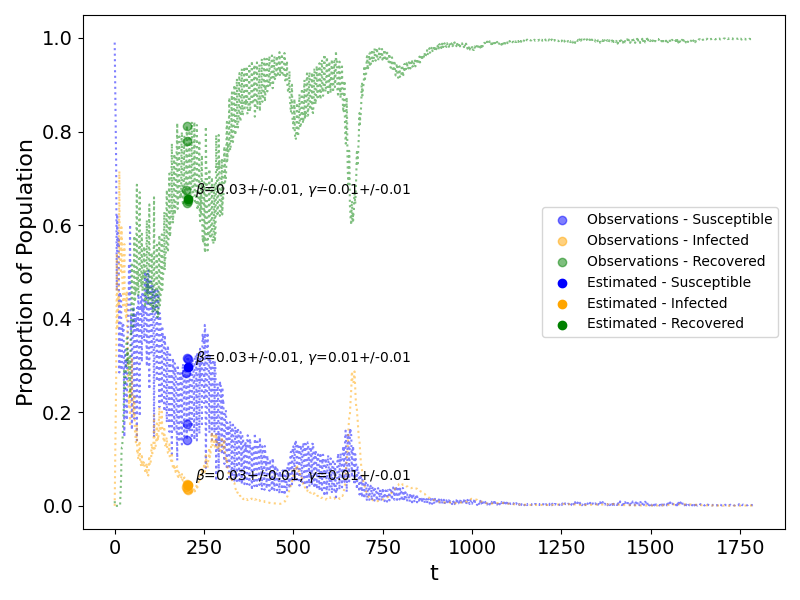}
        \caption*{\textbf{(a)}}
        \label{fig:multi_traj_SANFRANCOVID19_state_estimate_forward}
    \end{minipage}
    \hfill
    \begin{minipage}[b]{0.75\linewidth}
        \centering
        \includegraphics[width=\linewidth]{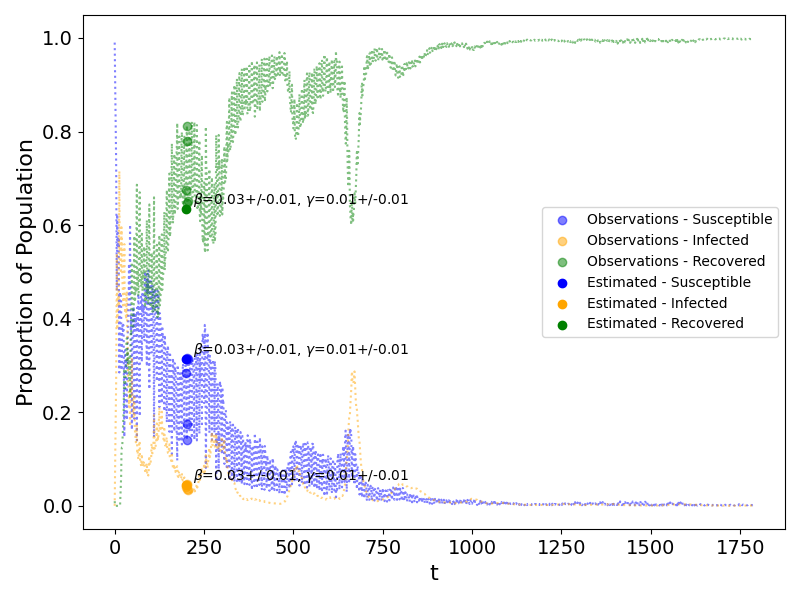}
        \caption*{\textbf{(b)}}
        \label{fig:multi_traj_SANFRANCOVID19_state_estimate_inverse}
    \end{minipage}
    \caption{Joint state and parameter estimation using normalizing flow with transformer conditioning operator trained on multiple trajectories with San Francisco COVID-19 dataset. (a) Forward state estimation. (b) Backward state estimation. }
    % We also add a visualization in gray for the SIR system's trajectories that corresponds to the mean of the estimated parameters.
    \label{fig:combined_SANFRANCOVID19_state_estimates}
\end{figure}

An additional feature that complements state prediction, and is particularly useful for applications in epidemiology is the ability to simultaneously estimate meaningful system parameters.
For the selected application, the transmission rate parameter $\beta$ and the recovery rate parameter $\gamma$ are key for domain experts to understand and classify the underlying disease and to inform public policy.
The proposed NF approach \rev{trained on multiple trajectories}, can be easily extended to explicitly estimate the joint density of the SIR states and parameters of the underlying system. 
An example of joint estimate of states and parameters for the San Francisco COVID-19 dataset is shown in Figure~\ref{fig:combined_SANFRANCOVID19_state_estimates} which reports the estimated parameters and quantifies their uncertainty. 
% and shows the trajectories of the SIR systems that corresponds to the mean estimated parameters.

\rev{To further show the contribution of this method, we use the state and parameter estimates from the proposed NF approach as inputs to a SMC algorithm. 
Specifically, we predict the mean underlying system state and parameters, $\beta$ and $\gamma$, using NF with five-observation context from the San Francisco COVID-19 dataset. 
Then, we use those predictions as inputs to SMC considering SIR model dynamics, and using real COVID-19 data as observations.
This hybrid method combines the capabilities of traditional filtering methods, which require known system parameters to be prescribed, with NF-based state and parameter estimation.
Results from SMC are shown in Figure~\ref{fig:pf_nf_hybrid}.
Comparing Figure~\ref{fig:pf_nf_hybrid} to the NF forward rollout predictions in Figure~\ref{fig:combined_SANFRANCOVID19_rollouts_transformer_fwd}, we observe that the NF rollout exhibits similar, if not greater, capabilities in predicting the underlying system than the traditional filtering method.
}

\rev{
\begin{figure}[!ht]
    \centering
    \begin{minipage}[b]{0.95\linewidth}
        \centering
        \includegraphics[width=\linewidth]{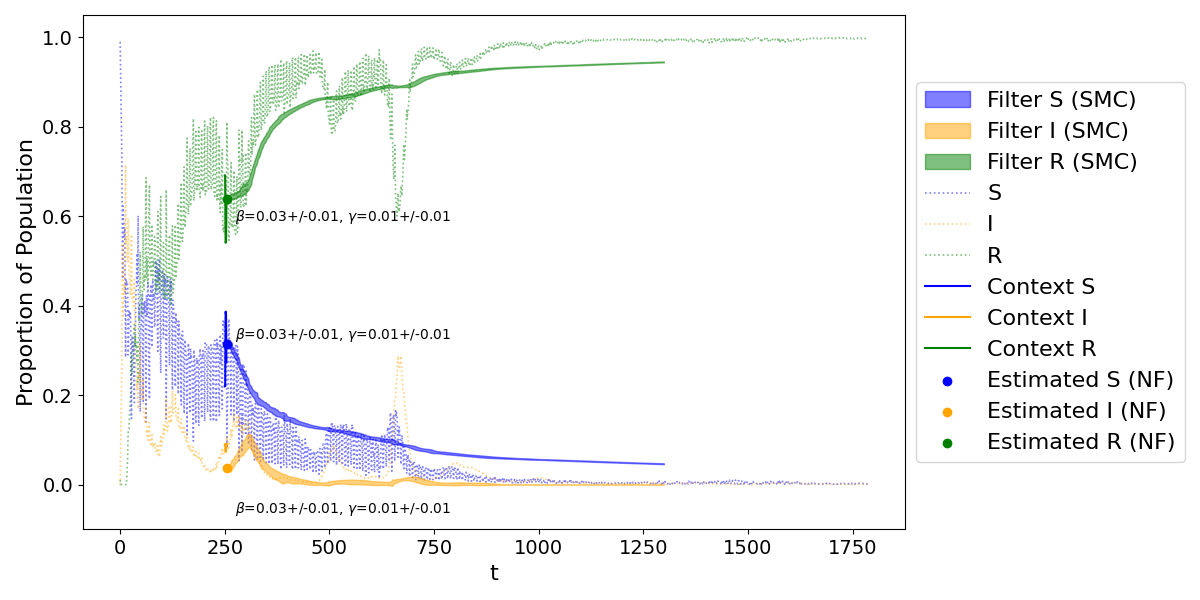}
        \caption*{\textbf{(a)} Full view}
        \label{fig:pf_nf_hybrid_full}
    \end{minipage}
    \hfill
    \begin{minipage}[b]{0.95\linewidth}
        \centering
        \includegraphics[width=\linewidth]{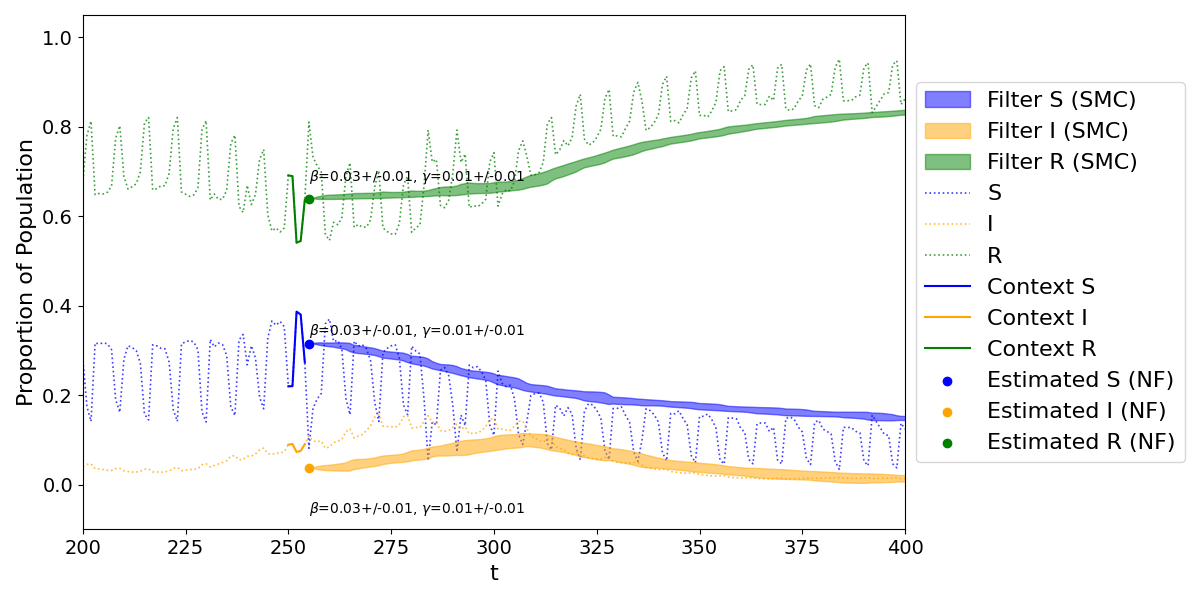}
        \caption*{\textbf{(b)} Zoomed}
        \label{fig:pf_nf_hybrid_zoomed}
    \end{minipage}
    \caption{$3\sigma$ filter using SMC (particle filter) with input state and parameters estimated from joint state and parameter normalizing flow.}
    \label{fig:pf_nf_hybrid}
\end{figure}
}

% ========================================================
\section{Discussion and Future Work}\label{sec:discussion}
% ========================================================

In this study, we consider a flexible family of NF-based approaches for online estimation of joint parameters and states that can easily combine information from physics or equation-based models, and observations from real processes. 
In particular, we evaluate the performance of different  conditioning operators based on transformers and state-space model\rev{s}, respectively, and the effect of adding an optimal-transport-based kinetic energy loss term.
\rev{First, we illustrate the proposed advantages of NF-based conditional estimators versus traditional SMC. NF better captures vehicle positions with multi-modal dynamics, also providing a number of architectural advantages.}
Each NF \rev{variant} showed effectiveness for state estimation when applied to the autonomous vehicle dataset from Delecki et. al~\cite{delecki2023deepnormalizingflowsstate}, with state-space models outperforming state-of-the-art transformers in some cases.
The addition of a modified optimal transport-inspired kinetic loss term showed improved sampling efficiency by optimizing the probability path between each transformation in the flow.
Expanding on the performance capability of the transformer conditioning operator, the optimal transport-inspired kinetic loss term provided overall \rev{stronger performance in estimating states for the autonomous driving dataset at the bifurcation.} \rev{Additionally, the Mamba-SSM conditioning operator performed particularly well when applied to the synthetic SIR dataset.}

Furthermore, we explore the application of our normalizing flow architecture in an epidemiological context. 
Mathematical models in their exact form, if at best, provide a highly generalized picture of a real-world system. That said, such dynamics, often developed by domain experts, contain important information about the reality of the underlying system. 
Operating under the assumption that the COVID-19 pandemic developed in a way somewhat described by the standard Susceptible-Infected-Recovered (SIR) model, we employ a conditional normalizing flow architecture to capture the underlying distribution in simulated noisy, multi-modal SIR trajectories and predict the next and previous states of real COVID-19 pandemic data.
The expressivity of the conditional NF accurately models the noisy system of multiple SIR trajectories and showed high accuracy in forward and backward state predictions. 
%from a noisy context selected from one of the many simulated trajectories. 
%
Despite having only been trained on synthetic noisy data, conditional NF rollout correctly captures forward and backward trends in realistic COVID-19 traces acquired by the San Francisco Department of Public Health.

% Conditional normalizing flows present a powerful method for capturing real-world system behavior that resembles an underlying mathematical model, combining deep learning methodology with domain expert research. 

Furthermore, the flexibility of the architecture allowed for the addition of system parameters as targets in training, allowing for the prediction of such parameters given a context of states in time. This allows for a more practical understanding of the underlying system dynamics. 
\rev{Given this flexibility, we were able to use estimated state and parameters as inputs to a SMC algorithm for real COVID-19 SIR data. This hybrid approach presents an interesting combination of our methods and traditional filtering. 
Also, the performance of such filtering using the traditional SMC algorithm compared to the forward rollout predictions of the NF architecture indicates promise in the capabilities of our rollout method.}

Future work will focus on further testing the efficacy in the joint prediction of states and parameters in more complex mathematical systems, and in fusing information from models with varying degree of fidelity and observations.

% =======================
\section{Acknowledgments}
% =======================

The authors acknowledge support from NSF CAREER award \#1942662 (DES) and NSF CDS\&E award \#2104831 (DES), and from NIH grant \#1R01HL167516. High performance computing resources for this study were provided by the Center for Research Computing at the University of Notre Dame.

% ============
% BIBLIOGRAPHY
% ============

\bibliography{biblio.bib}

\end{document}

%% file: fig_3a.tex
\begin{tikzpicture}[
    node distance=2.5cm and 3.5cm,
    >=Stealth,
    box/.style={
        rectangle,
        rounded corners=8pt,
        minimum width=3.2cm,
        minimum height=2.2cm,
        align=center,
        font=\normalsize\bfseries,
        text=textcolor,
        draw,
        line width=2pt,
        fill=white,
        blur shadow={shadow blur steps=5, shadow xshift=0pt, shadow yshift=-2pt, shadow blur radius=4pt}
    },
    smallbox/.style={
        rectangle,
        rounded corners=6pt,
        minimum width=3.2cm,
        minimum height=1.6cm,
        align=center,
        font=\normalsize\bfseries,
        text=textcolor,
        draw,
        line width=2pt,
        fill=white,
        blur shadow={shadow blur steps=5, shadow xshift=0pt, shadow yshift=-2pt, shadow blur radius=4pt}
    },
    label/.style={
        font=\small,
        text=textcolor
    },
    arrow/.style={
        ->,
        thick,
        color=arrowcolor,
        line width=1.2pt
    },
    flowlabel/.style={
        font=\normalsize,
        text=textcolor
    }
]

% Nodes
\node[smallbox, draw=basecolor] (base) {Base\\Distribution};
\node[smallbox, draw=targetcolor, right=of base] (target) {Target\\Distribution};
\node[smallbox, draw=transformercolor, above=1cm of base, xshift=3.35cm] (transformer) {Transformer\\Network};

% Input arrow to transformer (from above)
\draw[arrow] ([yshift=0.5cm]transformer.north) node[label, left, xshift=-2pt] {$\mathbf{o}_{1:t}$} 
    node[label, right, xshift=2pt] {Input vector} -- (transformer.north);

% Output arrow from transformer to flow
\draw[arrow] (transformer.south) -- node[label, left, xshift=-2pt] {$\mathbf{h}$} 
    node[label, right, xshift=2pt] {Parametrized vector} 
    ([yshift=0.8cm]$(base.east)!0.5!(target.west)$);

% Main flow arrow
\draw[arrow] (base.east) -- node[flowlabel, above] {Normalizing flow} 
    node[flowlabel, below] {$f = f_{1} \circ \ldots \circ f_{L}$} (target.west);

\end{tikzpicture}

%% file: fig_3b.tex
\begin{tikzpicture}[
    node distance=0.8cm,
    >=Stealth,
    % Style for vector blocks (sequences of cells)
    vecblock/.style={
        rectangle,
        minimum width=0.5cm,
        minimum height=0.5cm,
        draw=black,
        line width=0.8pt,
        fill=#1
    },
    % Style for operation boxes
    opbox/.style={
        rectangle,
        rounded corners=3pt,
        minimum width=2.5cm,
        minimum height=0.8cm,
        align=center,
        font=\small\bfseries,
        text=textcolor,
        draw=black,
        line width=1.5pt,
        fill=white
    },
    % Style for the main Mamba-SSM block
    mambabox/.style={
        rectangle,
        rounded corners=8pt,
        minimum width=2.8cm,
        minimum height=1.2cm,
        align=center,
        font=\normalsize\bfseries,
        text=black,
        fill=white,
        draw=black,
        line width=2pt
    },
    % Style for input boxes
    inputbox/.style={
        rectangle,
        minimum width=1.2cm,
        minimum height=0.8cm,
        align=center,
        font=\small\bfseries,
        text=textcolor,
        draw=black,
        line width=1pt,
        fill=inputcolor!30
    },
    % Style for dotted circles (activation functions)
    activation/.style={
        circle,
        minimum size=1.2cm,
        align=center,
        font=\scriptsize\bfseries,
        text=textcolor,
        draw=black,
        line width=1pt,
        dotted,
        fill=white
    },
    % Style for small symbols
    symbol/.style={
        circle,
        minimum size=0.6cm,
        align=center,
        font=\footnotesize,
        text=textcolor,
        draw=black,
        line width=1pt,
        dotted,
        fill=white
    },
    arrow/.style={
        ->,
        thick,
        color=arrowcolor,
        line width=1.2pt
    },
    label/.style={
        font=\small,
        text=textcolor
    }
]

% Input vectors at top (two copies)
\node at (0,0) (topleft) {};
\foreach \i in {1,...,3} {
    \node[vecblock=inputcolor!30] (topl\i) at ($(topleft)+(\i*0.6,0)$) {};
}

% Left branch - Upper projection and vector
\node[below=0.8cm of topl2, xshift=-1.5cm] (proj1center) {};
\foreach \i in {1,...,3} {
    \node[vecblock=inputcolor!50] (upp\i) at ($(proj1center)+(\i*0.6-1.5,0)$) {};
}
\node[below=0.05cm of upp2.south, font=\small] {Projection};

\foreach \i in {1,...,5} {
    \node[vecblock=inputcolor!70] (upv\i) at ($(proj1center)+(0,-1.2)+(\i*0.6-2.1,0)$) {};
}

% Convolution box
\node[opbox, below=0.6cm of upv3] (conv) {Convolution};

% SiLU activation (left)
\node[activation, below=0.8cm of conv] (silu1) {SiLU};

% Selective SSM box
\node[opbox, below=0.8cm of silu1, minimum width=3cm] (ssm) {Selective SSM};

% Activation symbol below SSM
\node[symbol, below=0.4cm of ssm] (mult1) {$\times$};
\node[right=0.05cm of mult1, yshift=0.2cm, font=\small] {Activation};
\node[right=0.05cm of mult1, yshift=-0.15cm, font=\small] {or multiplication};

% Output projection (lower)
\node[below=0.5cm of mult1] (proj3center) {};
\foreach \i in {1,...,5} {
    \node[vecblock=outputcolor!70] (lowv\i) at ($(proj3center)+(\i*0.6-1.8,0)$) {};
}
\node[below=0.05cm of lowv3.south, font=\small] {Projection};

\foreach \i in {1,...,3} {
    \node[vecblock=outputcolor!50] (lowp\i) at ($(proj3center)+(0,-1.2)+(\i*0.6-1.2,0)$) {};
}

% Right branch - projection and activation
\node[right=3.5cm of proj1center] (proj2center) {};
\foreach \i in {1,...,3} {
    \node[vecblock=inputcolor!50] (rpp\i) at ($(proj2center)+(\i*0.6-1.5,0)$) {};
}
\node[below=0.05cm of rpp2.south, font=\small] {Projection};

\foreach \i in {1,...,5} {
    \node[vecblock=inputcolor!70] (rpv\i) at ($(proj2center)+(0,-1.2)+(\i*0.6-2.1,0)$) {};
}

% SiLU activation (right) - placed outside container
\node[activation, below=2.2cm of rpv3] (silu2) {SiLU};

\node[label, left=0.35cm of silu2] {Activation};

% Multiplication symbol
\node[symbol, right=2cm of lowp2] (mult2) {$+$};

% Left side - Mamba-SSM block and inputs
\node[mambabox, left=2.5cm of conv] (mamba) {Mamba-SSM};

\node[inputbox, above=0.6cm of mamba] (o1t) {$\mathbf{o}_{1:t}$};
\node[inputbox, right=0.2cm of o1t] (z1t) {$\mathbf{z}_{1:t-1}$};
\node[label, left=0.1cm of o1t.west, yshift=0.2cm] {Observations and};
\node[label, left=0.1cm of o1t.west, yshift=-0.15cm] {latent variables};

\node[inputbox, below=0.6cm of mamba, fill=outputcolor!50] (h1) {$\mathbf{h}_t$};
\node[label, left=0.1cm of h1.west, yshift=0.2cm] {Mamba-SSM};
\node[label, left=0.1cm of h1.west, yshift=-0.15cm] {embedding};

% Normalize box at bottom
\node[opbox, below=0.8cm of mult2] (norm) {Normalize};

% Final output vector
\foreach \i in {1,...,3} {
    \node[vecblock=outputcolor!50] (out\i) at ($(norm)+(0,-1.2)+(\i*0.6-1.2,0)$) {};
}

% Main container box (define early to reference later)
\node[draw=black, line width=2pt, rounded corners=5pt, fit=(upv1)(rpv5)(upp1)(mult2), inner sep=0.4cm] (container) {};

% Arrows
% Connect two top inputs with continuous arrow

% Dotted lines from top inputs to projections (connecting extremes)
\draw[arrow, thick] (topl2.south) -- (topl2.south |- upp3.east) -- (upp3.east);
\draw[arrow, thick] (topl2.south) -- (topl2.south |- rpp1.west) -- (rpp1.west);

% Projection to vectors (dotted)
\draw[dotted, thick] (upp1.south west) -- (upv1.north west);
\draw[dotted, thick] (upp3.south east) -- (upv5.north east);
\draw[dotted, thick] (rpp1.south west) -- (rpv1.north west);
\draw[dotted, thick] (rpp3.south east) -- (rpv5.north east);

% Convolution flow
\draw[arrow] (upv3.south) -- (conv.north);
\draw[arrow] (conv.south) -- (silu1.north);
\draw[arrow] (silu1.south) -- (ssm.north);

% Right branch to SSM
\draw[arrow] (rpv3.south) -- (silu2.north);
\draw[arrow] (silu2.south) -- (silu2.south |- ssm.east) -- (ssm.east);

% SSM to multiplication
\draw[arrow] (ssm.south) -- (mult1.north);
\draw[arrow] (mult1.south) -- (lowv3.north);

% Projection flow
\draw[dotted, thick] (lowv1.south west) -- (lowp1.north west);
\draw[dotted, thick] (lowv5.south east) -- (lowp3.north east);

% To multiplication symbol
\draw[arrow] (lowp3.east) -- (mult2.west);
\draw[arrow] (topl3.east) -- ([xshift=0.5cm]container.east |- topl3.east) -- ([xshift=0.5cm]container.east |- mult2.east) -- (mult2.east);

% To normalize
\draw[arrow] (mult2.south) -- (norm.north);
\draw[arrow] (norm.south) -- (out2.north);

% Mamba-SSM connections
\draw[arrow] (o1t.south) -- (mamba.north);
\draw[arrow] (z1t.south) -- (mamba.north);
\draw[thick] (mamba.east) -- (mamba.east -| container.west);
\draw[arrow] (mamba.south) -- (h1.north);

\end{tikzpicture}